\documentclass[sigconf, 9pt]{acmart}
\AtBeginDocument{%
  }

\copyrightyear{2026}
\acmYear{2026}
\setcopyright{cc}
\setcctype{by}
\acmConference[MobiSys '26]{The 24th Annual International Conference on Mobile Systems, Applications and Services}{June 21--25, 2026}{Cambridge, United Kingdom}
\acmBooktitle{The 24th Annual International Conference on Mobile Systems, Applications and Services (MobiSys '26), June 21--25, 2026, Cambridge, United Kingdom}
\acmDOI{10.1145/3745756.3809242}
\acmISBN{979-8-4007-2027-7/2026/06}

\usepackage{graphicx,enumitem} 

\usepackage{booktabs}
\usepackage{tabularx}
\usepackage{makecell}
\usepackage{enumitem}
\usepackage{multirow}

\begin{document}

\title{[Emerging Ideas] Artificial Tripartite Intelligence: A Bio-Inspired, Sensor-First Architecture for Physical AI}

\author{You Rim Choi}
\authornote{Both authors contributed equally to this research.}
\affiliation{%
  \institution{Seoul National University}
  \city{Seoul}
  \country{South Korea}
}
\email{yrchoi@snu.ac.kr}
\orcid{0000-0002-1068-7403}

\author{Subeom Park}
\authornotemark[1]
\affiliation{%
  \institution{Seoul National University}
  \city{Seoul}
  \country{South Korea}
  }
\email{sbpark7@snu.ac.kr}
\orcid{0009-0006-6885-1372}

\author{Hyung-Sin Kim}
\authornote{Corresponding author.}
\affiliation{%
  \institution{Seoul National University}
  \city{Seoul}
  \country{South Korea}
  }
\email{hyungkim@snu.ac.kr}
\orcid{0000-0001-8605-5077}



\begin{abstract}
As AI moves from data centers to robots and wearables, scaling ever-larger models becomes insufficient. Physical AI operates under tight latency, energy, privacy, and reliability constraints, and its performance depends not only on model capacity but also on how signals are acquired through controllable sensors in dynamic environments. 
We present Artificial Tripartite Intelligence (ATI), a bio-inspired, sensor-first architectural contract for physical AI. ATI is tripartite at the systems level: a Brainstem (L1) provides reflexive safety and signal-integrity control, a Cerebellum (L2) performs continuous sensor calibration, and a Cerebral Inference Subsystem spanning L3/L4 supports routine skill selection and execution, coordination, and deep reasoning.
This modular organization allows sensor control, adaptive sensing, edge-cloud execution, and foundation model reasoning to co-evolve within one closed-loop architecture, while keeping time-critical sensing and control on device and invoking higher-level inference only when needed. 
We instantiate ATI in a mobile camera prototype under dynamic lighting and motion. In our routed evaluation (L3-L4 split inference), compared to the default auto-exposure setting, ATI (L1/L2 adaptive sensing) improves end-to-end accuracy from 53.8\% to 88\% while reducing remote L4 invocations by 43.3\%. These results show the value of co-designing sensing and inference for embodied AI.

\end{abstract}






\maketitle

\section{Introduction}
\label{sec:intro}

\begin{figure}[tbp]
    \centering
    \includegraphics[width=1\linewidth]{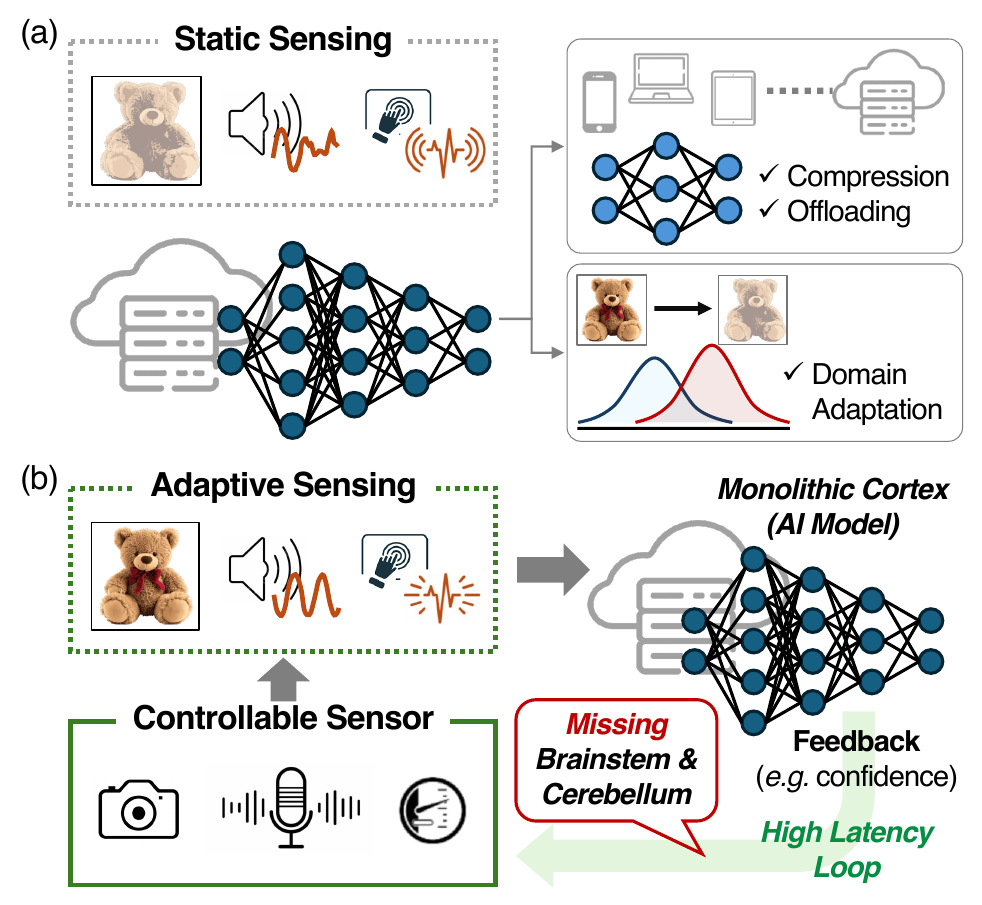}
    \caption{Limitations of prevailing paradigms for physical AI.
    (a) Computation-centric approach: sensors are treated as passive inputs while adaptation is confined to the model stack.
    (b) Cortex-centric adaptive sensing: sensing is controlled directly by a monolithic model without structured reflex and calibration layers.}
    \Description{Two-panel schematic comparing physical AI paradigms: computation-centric versus cortex-centric adaptive sensing.}
    \label{fig:prevailing_paradigms}
\end{figure}

The dominant recipe in modern AI has been to scale data, models, and cloud computation~\cite{brown2020language, vaswani2017attention, radford2021learning}. 
That recipe has delivered remarkable progress, but it does not transfer cleanly to \emph{physical, embodied AI}~\cite{hu2023toward,ma2024survey,liu2025aligning}.
In physical settings, inputs come from \textit{on-device sensors} (e.g., cameras, microphones, and IMUs) whose operating parameters are controllable and whose outputs are shaped by motion, lighting, occlusion, interference, and other environmental dynamics. These sensors must operate under the same tight latency, energy, privacy, and reliability constraints as the on-device and edge computation that they feed. In such systems, \textit{sensing is not merely a front-end to inference; it is part of the inference loop}. 
Yet most existing approaches remain computation-centric: they treat sensing as fixed and push adaptation downstream into model compression, domain adaptation, or edge-cloud offloading (Figure~\ref{fig:prevailing_paradigms}(a)).

\begin{table*}[t]
\centering
\caption{Adaptive human sensing versus conventional artificial sensors across four modalities.}
\label{tab:sensing}
\resizebox{1\textwidth}{!}{
\begin{tabular}{l l l}
\toprule
\textbf{Modality} & \textbf{Human sensor (adaptive)} & \textbf{Artificial sensor (static)} \\
\midrule
\textbf{Vision} &
\makecell[l]{
Pupil dynamically controls light intake.\\
Photoreceptors adapt to luminance over seconds.\\
Fast saccades redirect the fovea before cortical processing.}
&
\makecell[l]{
\emph{Camera} (controls: exposure, ISO, frame rate, focus).\\
Frame rate and exposure/ISO set in coarse steps, rarely changed.\\
Saturation and motion blur corrected mainly after capture.}
\\
\midrule
\textbf{Hearing} &
\makecell[l]{
Active gain control via outer hair cells over a wide dynamic range.\\
Efferent feedback protects against loud sounds.\\
Sub-millisecond binaural timing enables precise localization.}
&
\makecell[l]{
\emph{MEMS microphone} (controls: gain/AGC, sampling rate, beam pattern).\\
AGC, sampling rate, and bit depth fixed at design time.\\
3D localization relies on multi-mic arrays and heavy DSP.}
\\
\midrule
\textbf{Touch} &
\makecell[l]{
Dense, compliant skin with high spatial and temporal acuity.\\
Reflex arcs withdraw from painful stimuli before awareness.\\
Active exploration (rubbing) increases signal-to-noise ratio.}
&
\makecell[l]{
\emph{Tactile array / pressure sensor} (controls: readout freq, active ROI).\\ 
Sparse capacitive or piezoresistive taxel arrays on rigid surfaces.\\
Thresholds and sampling rates fixed; protection left to high-level control.}
\\
\midrule
\textbf{Proprioception} &
\makecell[l]{
Vestibular and proprioceptive organs track body pose and acceleration.\\
Rapid reflexes stabilize posture and gait without cortical involvement.}
&
\makecell[l]{
\emph{IMU} (controls: measurement range, sampling rate, filter parameters).\\
Accel/gyro data sampled at fixed rates with offline bias calibration.\\
Ranges and filters rarely change after deployment.}
\\
\bottomrule
\end{tabular}}
\end{table*}

Biological perception is organized differently. Before signals reach higher-level cognition, low-level mechanisms such as pupil reflexes and retinal gain control continuously regulate what is sensed and how it is encoded (Table~\ref{tab:sensing}). 
Recent work on adaptive sensing~\cite{lee2024learning, baekadaptive,baek2024unexplored,baekposition,im2025tokens} has begun to treat sensing as a control problem, but two limitations remain. Some approaches rely on heuristics that are only loosely coupled to downstream inference~\cite{lee2024learning}. Other approaches tightly couple sensing decisions to a downstream model, repeatedly querying it to maximize confidence~\cite{baekadaptive}. 
These efforts improve individual components, but they do not provide a general systems architecture for integrating sensing, fast control, and inference into a single closed loop (Figure~\ref{fig:prevailing_paradigms}(b)).

We argue that the missing piece is \textit{architectural}, not merely algorithmic or a matter of more compute. The history of computing suggests why: major inflection points often come from architectural contracts that define how components interact, not just from faster components. The von Neumann architecture, for example, changed computing not by simply adding more vacuum tubes or faster arithmetic units, but by establishing a \textit{systems contract} for programmable computation~\cite{von1993first}. 
Physical AI now needs a comparable contract for sensor-integrated intelligence: a \textit{sensor-first design} in which sensing and inference co-adapt under explicit latency, energy, and reliability constraints, rather than a one-way pipeline from sensor to model. In this view, inference should shape how the system senses, and sensing quality should in turn determine when more expensive computation is warranted.

To address this need, we propose \textbf{Artificial Tripartite Intelligence (ATI)}, a sensor-first architectural contract for physical AI. ATI is not a single model or control policy; it is a modular systems abstraction for jointly composing adaptive sensing and inference across multiple time scales.
ATI is \textit{tripartite at the systems level}: it separates intelligence into three functional parts---(i) Brainstem (L1), which provides reflexive safety and signal-integrity control; (ii) Cerebellum (L2), which continuously calibrates sensing; and (iii) a Cerebral Inference Subsystem spanning L3 and L4, which supports routine skill selection and execution, coordination, and deep reasoning. Together, these parts form a unified closed-loop perception stack in which sensor control and inference are tightly coupled.

At the implementation level, these three parts are realized as four layers plus a coordination mechanism: 

\begin{itemize}[leftmargin=*, nosep]
    \item \textbf{L1: Reflex control (Brainstem).} An ultra-fast layer at the sensor-actuator boundary that enforces safety constraints and protects signal quality.
    
    \item \textbf{L2: Sensor calibration (Cerebellum).} A sensor-rate adaptive layer that continuously adjusts sensing parameters such as exposure, gain, and region of interest (ROI) to maintain input quality.
    
    \item \textbf{L3: Routine skill selection/execution (Basal Ganglia Network, BGN).} An on-device layer that handles routine cases by selecting and executing the appropriate task-specific local model.
    
    \item \textbf{L4: Deep reasoning (Hippocampal--Cortical Network, HCN).} A higher-capacity layer, typically on the edge or in the cloud, that is invoked selectively for difficult or ambiguous cases.
    
    \item \textbf{L3--L4 coordination (Frontoparietal Network, FPN).} A control mechanism that routes between local execution and deep reasoning based on expected utility and system cost. This coordination belongs to ATI's third part, the Cerebral Inference Subsystem, rather than forming a separate part.
\end{itemize}

\noindent
ATI's novelty is therefore architectural rather than component-level. Although L3/L4 partially overlap with familiar fast/slow, or System 1/System 2 architectures for physical AI~\cite{bjorck2025gr00t}, ATI is broader and sensor-first: it elevates reflex control and continuous calibration (L1/L2) to \textit{first-class} system layers beneath inference, and it makes escalation from L3 to L4 an explicit resource-aware systems decision. By making reflex control, continuous calibration, local execution, and selective escalation explicit roles with distinct time scales and interfaces, ATI provides a modular structure in which sensor control, model adaptation, edge–cloud execution, and foundation models can co-evolve synergistically instead of as isolated optimizations.

\subsection{Why MobiSys, and why now?}
MobiSys has long explored efficient sensing, adaptation, and execution under tight device constraints~\cite{miluzzo2008sensing,cuervo2010maui,yi2020heimdall}. In parallel, modern AI has developed increasingly capable models that remain largely decoupled from real-time sensor control~\cite{radford2021learning,driess2023palm}. 
As these threads converge in embodied AI, the central systems question is no longer only how to run models efficiently, but how to \textit{co-design} sensing and inference as an end-to-end system. ATI offers a principled blueprint for that co-design.

\subsection{Contributions}
%

\begin{itemize}[leftmargin=*, nosep]
    \item \textbf{Sensor-first architecture.} 
    We identify why computation-centric designs are insufficient for physical AI and articulate a sensor-first perspective where sensing and inference form a closed loop. 

    \item \textbf{The ATI contract.} 
    We propose ATI, a tripartite, modular architectural contract that separates reflex control (L1), sensor calibration (L2), and an inference subsystem in which L3 handles routine skill selection and execution and L4 handles deep reasoning, while making coordination explicit. 

    \item \textbf{Prototype and outlook.} 
    We instantiate ATI in a vision-based mobile prototype, illustrating how active sensor tuning within ATI can improve robustness even with lightweight local models, and we discuss how ATI can serve as a common substrate for future advances in sensing and inference. 
\end{itemize}
\section{Related Work}
\label{sec:relatedwork}

ATI sits at the intersection of four lines of work: foundation models, mobile/edge execution, adaptive sensing and sensor control, and robotics middleware. These areas largely strengthen different parts of the stack, including high-capacity reasoning, efficient execution, sensor adaptation, and software infrastructure, whereas ATI provides the architectural framework that connects them.

\subsection{Foundation Models in Embodied/Physical AI}

Foundation models now provide high-capacity perception, semantic reasoning, and action generation for embodied/physical AI. 
CLIP~\cite{radford2021learning} established large-scale vision--language alignment, PaLM-E~\cite{driess2023palm} extended it to embodied settings by jointly encoding images, robot state, and language, and Vision--language--action (VLA) models, such as RT-2~\cite{zitkovich2023rt} and Gemini-based systems~\cite{team2025gemini}, further integrate perception and control.
In ATI, these models naturally map to L4, the high-capacity reasoning layer.

Recent robot foundation models have also adopted explicit fast and slow architectures. GR00T N1~\cite{bjorck2025gr00t} uses a tightly coupled dual-system in which System 2 performs deliberate reasoning and planning, while System 1 translates those plans into continuous robot actions. 
ATI partially overlaps with this fast/slow split in L3/L4, but differs in two important ways: (1) it elevates sensor-side reflex control and calibration (L1/L2) to first-class layers, and (2) it allows routine cases to be handled entirely at L3 without invoking L4.

\subsection{Computation-Centric Edge AI}

Mobile and edge sensing systems have primarily focused on \emph{where and how} inference should run under tight resource and latency constraints. 
Early work established hierarchical sensing~\cite{miluzzo2008sensing,lu2009soundsense} and computation offloading~\cite{cuervo2010maui,satyanarayanan2009case} to manage energy and latency.
With the rise of deep neural networks (DNNs), the emphasis shifted toward optimizing the model stack through compression~\cite{youn2022bitwidth,park2025distillsleep}, dynamic scaling~\cite{han2016mcdnn,fang2018nestdnn}, scheduling~\cite{chen2018marvel,yi2020eagleeye,jeong2022band,choi2022scriptpainter,kang2024mirror}, concurrent execution~\cite{yi2020heimdall,han2024pantheon,park2023pointsplit}, and portable acceleration (e.g., just-in-time kernel generation~\cite{jia2024empowering}).
These systems make computation placement and model efficiency first-class concerns, but they generally assume that the sensing pipeline is fixed. 
ATI differs by treating sensing itself as a first-class component and by explicitly structuring how sensing and inference interact across system layers.

\subsection{Adaptive Sensing and Sensor Control}

A complementary line of work treats sensing itself as part of the control loop. 
Classical systems adjust sensor parameters using hand-crafted heuristics, e.g., tuning exposure or gain to maintain target brightness.
More recent systems adopt learning-based \emph{adaptive sensing}. 
Several methods use reinforcement learning to control camera exposure and gain as part of a feedback loop, either to improve detection quality~\cite{lee2024learning} or to stabilize visual odometry under challenging lighting~\cite{zhang2024efficient}. 
Lens~\cite{baekadaptive} introduces a lightweight ``vision test'' module that evaluates candidate camera settings using model confidence as a quality proxy. 
Other frameworks control pan-tilt-zoom (PTZ) motion to keep low-confidence targets within a high-quality field of view~\cite{yang2024active}.
These systems show that sensing can be adapted to improve downstream performance. However, they are typically scoped to a single sensor and a specific task. 
ATI builds on this insight but addresses a broader architectural question: how should reflexive control, continuous calibration, skill selection/execution, and deep reasoning be organized into one closed-loop system?

\subsection{Robotics Middleware and Cloud Robotics}

Robotics middleware provides the software substrate on which embodied systems are built. Robot Operating System (ROS) and ROS~2 expose abstractions such as nodes, topics, and services for composing modular robot software~\cite{quigley2009ros,macenski2022robot}. 
Orocos~\cite{bruyninckx2001open} extends this model with stronger real-time control primitives.  
More recently, FogROS2~\cite{ichnowski2023fogros2} integrates ROS~2 with cloud and fog computing to support distributed execution, and fault-tolerant extensions  address the unreliability of remote infrastructure~\cite{chen2024fogros2,chen2025fogros2}.
ATI complements rather than replaces these platforms. ROS, Orocos, and FogROS2 provide mechanisms for composition, communication, and execution; ATI provides the architectural contract that organizes sensing, control, and inference roles across those mechanisms.
\section{Artificial Tripartite Intelligence (ATI)}
\label{sec:ati}

We describe ATI as a concrete systems architecture. We first distill the biological principles that motivate ATI, then explain how ATI's roles are placed across device, edge, and cloud resources, how they communicate through small interfaces, and how this organization supports efficient and robust perception in physical environments. We use vision as the running example because camera control and visual neuroscience provide clear reference points, but the same architectural principles extend to other sensing modalities.

\begin{table*}[t]
\centering
\small
\caption{Biological inspirations as conceptual motifs for ATI roles.}
\label{tab:ati_bio_mapping}
\resizebox{1\textwidth}{!}{
\begin{tabular}{p{0.27\linewidth} p{0.31\linewidth} p{0.34\linewidth}}
\toprule
\textbf{Biological inspiration} & \textbf{Functional role in biology} & \textbf{ATI component} \\
\midrule
Brainstem & Reflex and safety control & \textbf{L1} safety reflex and signal protection \\
Cerebellum & Predictive calibration and error correction & \textbf{L2} sensing calibration and sensor control \\
Basal Ganglia Network (BGN) & Skill selection and habitual execution & \textbf{L3} task-specific skill selection/execution on device \\
Frontoparietal Network (FPN) & Flexible coordination across systems & \textbf{Routing} (L3--L4 coordination) \\
Hippocampal--Cortical Network (HCN) & Memory and relational reasoning & \textbf{L4} deep reasoning and knowledge-based inference \\
\bottomrule
\end{tabular}}
\end{table*}

\begin{figure*}[tbp]
    \centering
    \includegraphics[width=\linewidth]{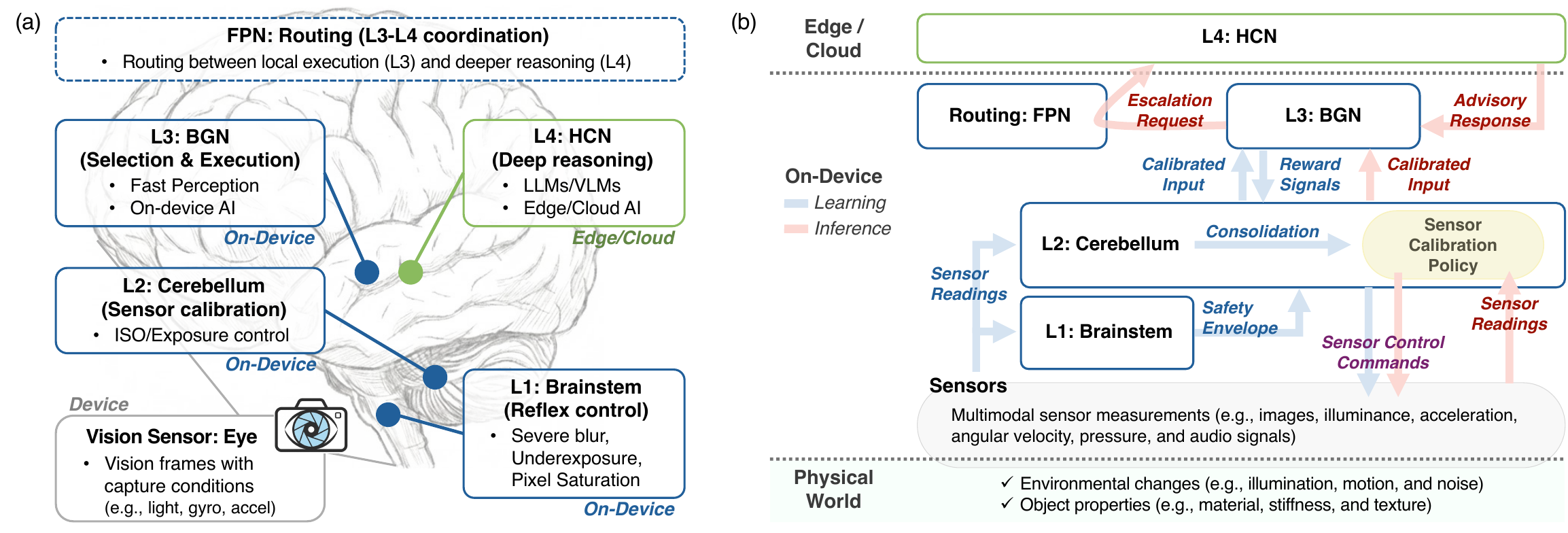}
    \caption{ATI architecture. (a) ATI role mapping for a camera-based system. (b) General ATI interfaces, from physical-world sensor inputs to on-device processing and edge/cloud reasoning. 
    }
    \Description{ATI architecture overview: role mapping for a camera-based system and general interfaces from sensing to on-device and edge or cloud reasoning.}
    \label{fig:ati_overview}
\end{figure*}

\subsection{Biological Inspirations and System Mapping}

ATI begins from a simple systems observation: in physical environments, many failures \textit{originate at capture} and cannot be repaired downstream. Saturation, motion blur, and low signal-to-noise ratio arise during acquisition. Once information is lost at that stage, larger models may partially compensate for it, but they cannot reliably recover it or guarantee timely, safe responses under tight latency budgets. 
This distinguishes embodied AI from the setting in which end-to-end deep learning has been most successful. When inputs are stable and pre-captured, much of the complexity can be absorbed into a single inference stack. In embodied systems, by contrast, the system must decide not only how to interpret inputs, but also how to \textit{acquire} them in changing environments.

Biological perception addresses this problem through \textit{layered control} rather than a  monolithic processor. ATI draws three system-level lessons from this organization.
First, fast reflex pathways protect signal integrity before higher-level reasoning. Pupillary reflexes regulate incoming light to prevent saturation, and vestibulo-ocular reflexes (VOR) stabilize retinal input during motion~\cite{ito1998cerebellar,do2019melanopsin}. They  operate on millisecond time scales and act before cortical processing, ensuring that downstream stages receive usable signals. 
ATI captures this principle in the Brainstem (L1), which enforces safety and signal-integrity constraints at the sensor interface.

Second, the cerebellum provides predictive, continuous calibration.
Unlike simple reflexes, it coordinates longer control loops such as smooth pursuit (stabilizing tracking)~\cite{lisberger2015visual} and saccade adaptation (calibrating gaze shifts)~\cite{hopp2004characteristics}. Through learned \textit{internal models}~\cite{ito2008control}, it predicts how actions affect future sensory states and corrects errors over time, stabilizing perception under dynamic conditions. 
ATI maps this principle to the Cerebellum (L2), which continuously adjusts sensing parameters to keep inputs within a stable, model-friendly operating regime.

Third, higher-level perception is functionally divided between fast routine processing and slower, context-dependent reasoning.
Biological systems combine rapid feed-forward responses for common situations with slower recurrent processing that incorporates memory, context, and relational structure when the scene is ambiguous or complex~\cite{thorpe1996speed,lamme2000distinct}. 
ATI reflects this division within the Cerebral Inference Subsystem by separating fast routine skill selection and execution in the Basal Ganglia Network (L3; BGN) from deeper reasoning in the Hippocampal-Cortical Network (L4; HCN), with the Frontoparietal Network (FPN) governing when routine processing is sufficient and when computation should escalate~\cite{mink1996basal, mink2018basal, mcclelland1995there, cole2013multi, duncan2010multiple}.
This organization allows L3 to execute quickly on-device, while L4 leverages edge or cloud resources for slower but more complex inference, supporting efficient operation in resource-constrained systems.

Table~\ref{tab:ati_bio_mapping} summarizes the mapping from these biological inspirations to ATI components. These correspondences are intended as design abstractions rather than literal anatomical equivalences.

\subsection{Sensor-First Roles and Placement}

Operationally, ATI's three system-level parts are realized as four
layers plus one coordination mechanism, as illustrated in Figure~\ref{fig:ati_overview}. Each role is defined by its
system responsibility, latency budget, and hardware placement across device, edge, and cloud resources. Together, these roles form a layered control system in which sensing, routine local selection/execution,  coordination, and deep reasoning are separated for clarity but coupled through explicit interfaces.

\noindent\textbf{Reflex Control (L1, Brainstem; on-device).} 
L1 is a reflex plane tightly integrated with the sensor driver and actuators, enforced by design to maintain  signal integrity and safety regardless of higher-level decisions. It operates within a sub-frame latency budget ($<$10\,ms) and enforces hard constraints such as anti-flicker guards, anti-saturation limits, and hardware safety bounds, preventing severely degraded/unsafe inputs from propagating downstream.

\noindent\textbf{Sensor Calibration (L2, Cerebellum; on-device).} 
L2 is an adaptive control plane operating at sensor rate. It continuously adjusts sensing parameters such as exposure, gain, tone mapping, and region of interest (ROI) in order to improve downstream task performance while remaining within L1's safety envelope. To do so, it uses task feedback from higher layers to keep inputs in a task-appropriate, model-friendly regime, while also relying on lightweight input-quality signals such as blur, saturation, and signal-to-noise ratio (SNR) for stable adaptation.

\noindent\textbf{Routine Skill Selection and Execution (L3, BGN; on-device).}
L3 is a lightweight perception and control module running on the device accelerator (e.g., NPU or GPU). 
Under a strict latency budget, it handles common cases by selecting the appropriate task-specific local model and executing it, producing outputs such as presence decisions, ROIs, short-term tracks, and coarse semantic outputs.
It also estimates calibrated uncertainty and interacts with the coordination mechanism to determine whether local handling is sufficient or whether the input should be escalated. Because L3 remains on device, basic interactivity continues even under poor or intermittent connectivity.

\noindent\textbf{Deep Reasoning (L4, HCN; edge/cloud).} 
L4 is a high-capacity reasoning module deployed on edge/cloud resources. It addresses complex or long-tail tasks such as fine-grained recognition, OCR, pose estimation, and multimodal reasoning. Due to varying network conditions, L4 is advisory and deadline-bounded: it can refine local predictions, but never blocks real-time control or safety loops.

\noindent\textbf{L3--L4 Coordination (FPN).} 
The coordination mechanism mediates between L3 and L4. It makes resource-aware routing decisions based on uncertainty, sensing quality, and latency constraints, deciding whether an input should be processed locally or escalated for deeper reasoning. By separating coordination from both local routine execution and deep reasoning, ATI makes cross-layer interaction explicit rather than implicit.

\subsection{Minimal Interfaces: The Contract}

To keep ATI portable and debuggable, each role interacts only through small, auditable interfaces as follows:

\noindent\textbf{Sensor Control Interface (L1/L2 $\to$ Sensor).} 
L1 and L2 share a low-level interface to the sensor and local actuators, but with different authority. L1 defines hard safety envelopes, and L2 operates only within those bounds. Table~\ref{tab:apis} lists representative primitives exposed by this interface. 
L1 uses these primitives to enforce caps and safe operating sets (e.g., rejecting shutter settings that induce flicker or clamping exposure to avoid thermal stress). Within those bounds, L2 performs fine-grained per-frame adjustments at video rate to keep capture within a model-friendly regime.

\begin{table}[t]
    \centering
    \caption{Example sensor control APIs exposed by ATI for camera and gimbal control.}
    \label{tab:apis}
    \footnotesize 
    \resizebox{\columnwidth}{!}{%
        \begin{tabular}{l l}
            \toprule
            \textbf{Function} & \textbf{Description} \\
            \midrule
            \texttt{set\_shutter(step)} & Exposure duration (discrete shutter step) \\
            \texttt{set\_gain(dB)}      & Analog signal amplification (in dB) \\
            \texttt{set\_hdr(mode)}     & HDR / WDR bracketing strategy \\
            \texttt{set\_roi(box)}      & Active pixel region read-out (ROI) \\
            \texttt{slew\_gimbal($\Delta$p, $\Delta$t)} & Pan/Tilt actuation velocity \\
            \bottomrule
        \end{tabular}
    }
\end{table}

\noindent\textbf{Sensor-Execution Loop (L2 $\leftrightarrow$ L3).} 
This bidirectional interface couples active sensing with fast perception.
\begin{itemize}[leftmargin=*]
    \item \textbf{Downstream (L2 $\to$ L3):} For each frame, L2 attaches a quality vector (QV) containing metrics such as blur score, saturation ratio, and estimated lux. This allows L3 to distinguish between cases such as ``empty scene'' and ``blinded sensor'' and calibrate its uncertainty accordingly.
    \item \textbf{Upstream (L3 $\to$ L2):} L3 sends task-level feedback, such as detection confidence, target ROI, or a resample request. 
\end{itemize}
Importantly, the L2–L3 interaction is \textit{temporary} while L2 learns to operate in a new environment. Over time, L2 can \textit{internalize} L3's feedback into its own model.  Early in deployment, it may adjust parameters reactively, for example by raising gain when confidence drops. 
As learning progresses, however, L2 can maintain favorable capture settings \textit{proactively} through its internal model, without waiting for L3 to signal each failure case. This progression loosely mirrors how repeated practice in biological systems turns feedback-driven correction into rapid, automatic control.

\noindent\textbf{Escalation Interface (L3 $\to$ L4).} 
When L3 remains uncertain even after sensing has been optimized, it issues a targeted escalation request to L4 via
\texttt{escalate(payload, budget)}. The \texttt{payload} is chosen to minimize bandwidth (e.g., a cropped ROI rather than the full frame) and the response is advisory rather than blocking. 
To prevent stale updates, each response carries the originating frame timestamp so that L3 can discard results that arrive after the scene has changed. Importantly, L4 never controls the sensor directly. It interacts only with L3, which may translate L4's semantic feedback into new requests to L2 while preserving ATI's latency hierarchy.

\subsection{Quality-Aware Routing Under Latency Constraints}
\label{sec:QA_routing}

On each frame, ATI's coordination mechanism asks a simple question: \textit{is escalation to L4 worth it given the current uncertainty, input quality, and resource budget?} 
The decision uses four signals: 
(1) calibrated uncertainty $u(x)$ from L3, 
(2) the quality vector (QV) from L2, 
(3) current resource state, including network round-trip time $T_{RTT}$ and device energy headroom, and 
(4) task metadata, including the deadline $T_{deadline}$ and the cost of error.

L4 is invoked only when all of the following conditions hold:
\begin{itemize}[leftmargin=*, nosep]
    \item \textbf{High uncertainty:} L3 is genuinely unsure ($u(x) > \tau_{task}$).
    \item \textbf{Sufficient input quality:} The QV indicates that the sensor is not effectively  blinded. If the frame remains severely degraded despite L2's best efforts, ATI does not offload it; instead, it requests a resample from L2.
    \item \textbf{Feasible deadline:} The remaining time budget can accommodate the full escalation cycle: $T_{now} + T_{RTT} + T_{inf\_L4} \le T_{deadline}$
    \item \textbf{Positive net benefit:} The expected reduction in uncertainty outweighs the estimated communication cost.
\end{itemize}
This makes offloading a quality-aware routing decision rather than a blind fallback by filtering uninformative inputs locally. 
Crucially, local control loops never stall waiting for L4. L4's output is a refinement; if it arrives too late, it is discarded.

\subsection{Role-Aligned Learning Strategy}

ATI does not rely on monolithic end-to-end training. Instead, each module is trained or configured according to its architectural role.

\begin{itemize}[leftmargin=*, nosep]
    \item \textbf{Reflex control (L1, Brainstem): Configured, not learned.}
    Safety constraints are engineered and validated via regression tests. They are treated as deterministic rules to guarantee non-bypassable safety. 
    
    \item \textbf{Sensor calibration (L2, Cerebellum): Task-driven policy.} 
    L2 is trained as a control policy whose objective is downstream task performance. Offline data can be used to learn mappings from scene conditions to sensor settings. During deployment, L2 performs bounded adaptation within L1's safety envelope using feedback from L3.

    \item \textbf{Routine skill selection and execution (L3, BGN): Distillation on stabilized inputs.} 
    Because L2 keeps inputs within a stable operating regime, L3 need not learn invariance to all sensor artifacts. Instead, it can be trained via distillation from L4, using hard examples identified during routing. 

    \item \textbf{Deep reasoning (L4, HCN): Generalist teacher.}
    L4 is a pre-trained foundation model that resolves difficult cases and provides supervision signals for both L3 and L2. 

    \item \textbf{L3--L4 coordination (FPN): Data-driven routing.} 
    It is calibrated from offline statistics over uncertainty, input quality, latency, and task outcomes. This yields routing boundaries that satisfy the benefit-cost trade-off in Section~\ref{sec:QA_routing}.
\end{itemize}
This role-aligned decomposition offers two system-level benefits.
First, it improves \textit{data efficiency}: because L2 stabilizes the input distribution, L3 can be trained on cleaner and narrower operating conditions. Second, it improves \textit{compute efficiency}: L1 and L2 act as pre-semantic filters, preventing wasted computation on inputs that are unsafe, irrecoverable, or low value.
\section{ATI Prototype: Lab-Based Object Classification on a Closed Track}
\label{sec:implementation}

\begin{figure}[t]
    \centering
    \includegraphics[width=\linewidth]{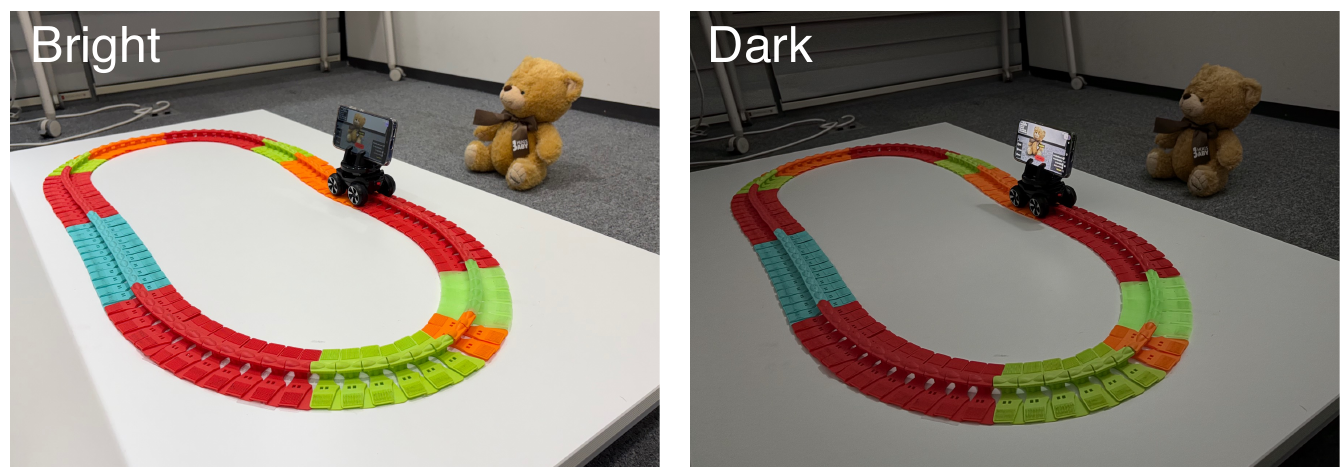}
    \caption{Experimental setup for the prototype. 
    A Galaxy S25 is mounted on a small vehicle that drives along a closed track. 
    Ambient illumination is set to bright (left) and dark (right).}
    \Description{Prototype setup with a smartphone-mounted vehicle on a closed track under bright and dark lighting conditions.}
    \label{fig:exp_setting}
\end{figure}

To demonstrate ATI's feasibility, we built a vision-based prototype (Figure~\ref{fig:exp_setting}) that runs the device-side layers (L1–L3) on a commodity smartphone and realizes the L4 path through a remote inference. This section describes the task scenario, hardware setup, and concrete implementation of each ATI role.

\subsection{Task Scenario}

Our testbed mounts a Galaxy S25 smartphone on a small car that repeatedly traverses a closed track. A custom Android application continuously captures RGB frames and runs the ATI pipeline over the live camera stream.

\noindent\textbf{Task Definition.}
The task is \textit{lap-based} object classification. In each trial, a single object is placed at a fixed location along the track. 
During each lap (about 3\,s), the object appears intermittently in the camera view under changing viewpoint, distance, and motion blur.
The system must identify the object class before the lap ends. The task stresses not only recognition accuracy, but also the robustness of the sensing stack under motion-induced image degradation.

\noindent\textbf{Exposure Control.}
We evaluate the system under two illumination regimes, ``bright'' and ``dark''. On a moving platform, camera control must balance light collection, motion blur, and sensor noise. Longer shutter times collect more light but increase motion blur, whereas higher ISO preserves short exposures at the cost of amplified noise. On the Galaxy S25, ATI directly controls these two camera parameters: exposure time (shutter speed) and ISO gain.

\noindent\textbf{Inference Path.} 
For local inference (the L3 path), the application supports several lightweight backbones, including MobileNetV2~\cite{sandler2018mobilenetv2}, MobileNetV3~\cite{howard2019searching}, and EfficientNet-Lite0~\cite{tan2019efficientnet}, all pre-trained on ImageNet~\cite{krizhevsky2012imagenet}. 
Because this prototype targets a single application, L3 does not need to select among multiple routine skills and therefore reduces in practice to a single routine local execution path: image classification.
To realize the deep reasoning path (L4), we send selected frames to a remote Gemini model~\cite{comanici2025gemini} in an offline replay setting. This lets us validate ATI's split-inference design without conflating it with live network variability.

\subsection{L1: Reflex Control and Safety Envelope}

L1 acts as a gatekeeper at the sensor layer to prevent unsafe or unusable inputs from propagating downstream. In the prototype, it computes a reflexive camera setting directly from motion and illumination measurements. Its control logic combines motion-aware exposure limiting and noise-constrained gain control, while enforcing a hard safety envelope that prevents catastrophic information loss under extreme low-light conditions.

\begin{itemize}[leftmargin=*, nosep]
    \item \textbf{Motion-based baseline exposure time.}
    L1 derives candidate exposure times $T_{safe}$ from both current accelerometer and gyroscope  measurements (i.e., $acc_t$ and $gyro_t$), mapping larger motion magnitudes to shorter exposures. It takes the smaller of the two, $\min(T_{safe}(acc_t), T_{safe}(gyro_t))$, as the raw motion-based target so that the stricter motion constraint is always enforced.

    \item \textbf{Illuminance-based safety floor.}
    While the motion-based target aggressively prevents blur, extreme low-light conditions (e.g., $<25$ lux) introduce the risk of near-complete signal loss. To counter this, L1 enforces a strict, illuminance-dependent minimum exposure time $T_{safe}(\ell_t)$, where $\ell_t$ denotes the current illuminance measured by the light sensor. 
    If the raw motion-based target falls below this threshold, L1 clamps it to $T_{safe}(\ell_t)$. This explicitly prioritizes structurally valid inputs over blur minimization, since downstream vision models can often tolerate moderate blur better than signal collapse.

    \item \textbf{ISO compensation with noise-aware limits.}
    L1 computes a baseline ISO ($ISO_t$) from the current illuminance ($\ell_t$) and exposure setting ($Exp_t$) to recover brightness. We define a reference exposure time $Exp_{def}=1/60$s based on the typical auto-exposure range for static indoor objects under normal daytime room lighting in our setup, and an illuminance-dependent quantity $ISO_{def}(\ell_t)$ using a calibrated mapping from the light sensor measurement under this reference exposure. L1 then adjusts ISO according to the ratio $Exp_{def}/Exp_t$, with shorter exposures requiring higher ISO and longer exposures allowing lower ISO. The resulting ISO is finally bounded by the noise-aware cap $ISO_{cap}(\ell_t)$ within hardware-viable ranges.
\end{itemize}

\noindent\textbf{Baseline Setting and Hardware Clamping.}
Formally, the exposure and ISO settings are computed as
\[
    Exp_t = \max\!\Big(\min\big(T_{safe}(acc_t), T_{safe}(gyro_t)\big),\ T_{safe}(\ell_t)\Big)
\]

\[
    ISO_t = \min\!\left( ISO_{def}(\ell_t)\cdot\frac{Exp_{def}}{Exp_t},\ ISO_{cap}(\ell_t) \right)
\]

\noindent
Any higher-level adjustment from L2 is constrained to remain within the safe envelope
\[
    \mathcal{S}_{safe,t} = \{(Exp, ISO)\mid Exp \ge T_{safe}(\ell_t),\ ISO \le ISO_{cap}(\ell_t)\}.
\]

\subsection{L2: Sensor Calibration via Contextual Bandits}

\begin{figure}[t]
    \centering
    \includegraphics[width=.79\columnwidth]{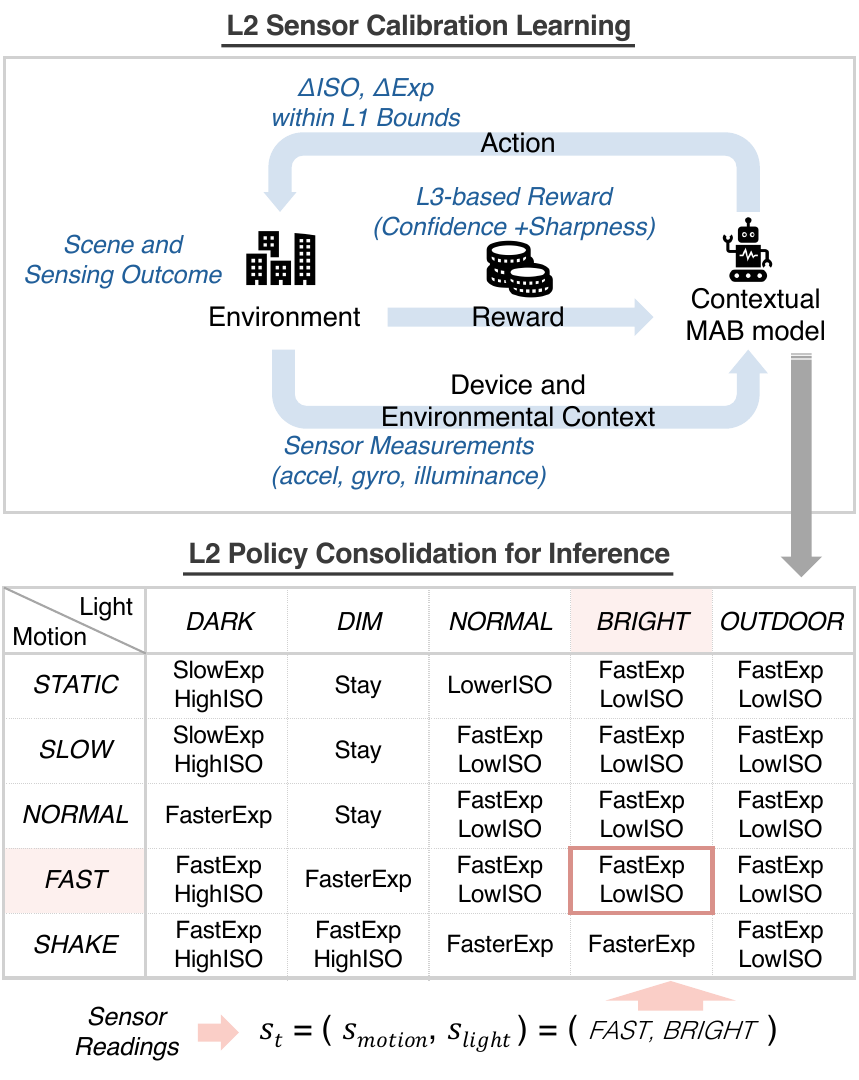} 
    \caption{L2 sensor calibration via contextual bandits: learning and consolidated policy lookup.}
    \Description{}
    \label{fig:l2_policy}
\end{figure}

Within L1's safety envelope, L2 performs fine-grained sensor calibration using feedback from L3. 
By associating camera settings with classification confidence, L2 learns an \textit{internal model} of which sensing strategy works best for a given motion and lighting context.
Figure~\ref{fig:l2_policy} summarizes this process, from contextual bandit learning to policy consolidation for inference.

\noindent\textbf{Formulation.} 
We implement L2 as a \textit{contextual multi-armed bandit (CMAB)}. 
This choice matches the structure of the prototype: camera actions (exposure time and ISO) affect immediate image quality, but they do not materially change the future environment state (ambient light/motion). 
L2 therefore focuses on selecting the best action for the current context rather than learning long-horizon state transitions.

\noindent\textbf{Context and Action.}
The \textit{context} $s_t$ is a discretized motion--light state used by the contextual bandit. 
Here, $s_{motion}$ denotes one of five motion states derived from accelerometer and gyroscope measurements, and $s_{light}$ denotes one of five illuminance states derived from the light sensor:
\[
    s_t = (s_{motion}, s_{light})
\]
This yields 25 discrete contexts.
In the prototype, L1 first computes a baseline setting $(Exp_t, ISO_t)$ from motion and illuminance, with the state determined using lap-averaged sensor values. This setting is mapped to the nearest supported exposure/ISO grid, after which L2 applies small relative offsets:
\[
    a_t = (\Delta \mathrm{Idx}_{ISO}, \Delta \mathrm{Idx}_{Exp}) \in \{-1, 0, +1\}^2
\]
Thus, L2 explores only a local neighborhood around L1's mapped baseline rather than the full parameter space.
The adjusted indices are clipped to valid hardware ranges before being applied, and any unsafe proposal is overridden by L1's safety envelope $\mathcal{S}_{safe,t}$.
Exploration therefore never compromises safety.

\noindent\textbf{Reward Signal.} 
Because the task objective is lap-based, the reward $r_t$ is computed once per lap rather than once per frame.
We define
\[
    r_t = \alpha \cdot \max(\mathrm{Conf}_{L3}) + (1-\alpha)\cdot \mathrm{Sharpness}
\]

\noindent
where $\max(\mathrm{Conf}_{L3})$ is the highest classification confidence produced by L3 during the lap.
The $\mathrm{Sharpness}$ score is evaluated on the frame that yielded this maximum confidence.
To penalize severe motion blur, we quantify sharpness using the variance of the Laplacian~\cite{pech2000diatom} applied to the downsampled grayscale image.
This variance is then normalized to a $[0,1]$ scale to yield the final $\mathrm{Sharpness}$ score.
We set $\alpha=0.9$ to prioritize classification confidence while still discouraging settings that result in heavy motion blur.

\noindent\textbf{Policy Consolidation.} 
To mimic biological motor learning, we implement a consolidation mechanism that converts repeatedly reinforced context--action associations in L2's \textit{internal model} into a persistent \textit{lookup table}. Over time, frequently recurring contexts can therefore be handled through internalized, automatic sensor responses in the sensing stack (L1/L2), reducing repeated dependence on feedback from L3.

\subsection{L3/L4/FPN: Split Inference with Lap-Level Coordination}

ATI's cerebral inference subsystem is instantiated as a split design with a fast on-device inference path (L3), a remote large-model path (L4), and a lap-level coordination policy (FPN).

\noindent\textbf{L3: Routine On-Device Skill Execution.}
L3 serves as the routine local skill for on-device image classification, enabling low-latency inference directly on the smartphone.
Unless otherwise noted, this skill is implemented with EfficientNet-Lite0.
For each frame, L3 outputs a predicted class and confidence score.

\noindent\textbf{L4: Remote Deep Inference.}
L4 is implemented with the Gemini API (\texttt{Gemini-2.5-flash-lite}~\cite{comanici2025gemini}).
A structured prompt asks the model to identify the main ImageNet-1K object and return JSON containing the class name and confidence.
Responses corresponding to non-ImageNet background content are filtered by setting their confidence to zero.
In this prototype, L4 is evaluated through offline replay of captured frames to avoid live network variability.

\noindent\textbf{FPN: Lap-Level Coordination and Routing.}
In the full ATI architecture, routing is driven primarily by local uncertainty and input quality, while also respecting latency and resource constraints (Section~\ref{sec:QA_routing}).
In this prototype, however, the task objective is to correctly classify  within each lap.
We therefore perform coordination at lap granularity, prioritizing accuracy while minimizing costly remote inference.
Empirically, the L4 path incurs about 70$\times$ higher latency than local L3 inference (mean: 32\,ms for on-device EfficientNet-Lite0 vs.\ 2,220\,ms round-trip for Gemini 2.5 Flash-Lite).

During each lap, the coordinator tracks the frame with the highest L3 confidence together with its sharpness score.
At lap completion, it applies the following policy:
\begin{itemize}[leftmargin=*, nosep]
    \item \textbf{Local acceptance:} If the peak L3 confidence exceeds the threshold $\tau_{conf}$, the system accepts the corresponding L3 prediction. If the peak confidence is below $\tau_{conf}$ but the sharpness score is also below the validity threshold $\tau_{valid}$, indicating severe motion blur, the system still accepts the L3 prediction rather than escalating. This avoids wasting remote inference on inputs that are too degraded to benefit from L4.
    \item \textbf{Escalation and conditional L4 acceptance:} If the peak confidence is below $\tau_{conf}$ and the sharpness score exceeds $\tau_{valid}$, the selected frame is escalated to L4. L4 returns its predicted label and confidence score, and the system accepts the L4 prediction only when its confidence exceeds that of L3. 
    This preserves the advantage of the specialized local path in cases where it remains more reliable than the more general remote model.
\end{itemize}
This policy reserves remote reasoning for sharp but ambiguous cases.
Confident local predictions are accepted locally, while severely blurred inputs are filtered before offloading.
As a result, L4 is invoked only when optimized local sensing (L1/L2) and routine local inference (L3) are still insufficient to produce a reliable decision.
\section{Evaluation}
\label{sec:evaluation}

\begin{table}[tb]
\centering
\caption{Evaluated configurations formed by combining sensing strategies and inference paths.}
\setlength{\tabcolsep}{3pt} 
    \resizebox{1\columnwidth}{!}{%
\begin{tabular}{llccc}
\toprule 
\textbf{Category} & \textbf{Control Method} & \textbf{Local-Only} & \textbf{Remote-Only}  & \textbf{Split/Routing} \\
\midrule 
Baseline & Default AE & AE-L3 & AE-L4& AE-L3-L4  \\
Baseline & Vendor EIS & EIS-L3 & EIS-L4 & EIS-L3-L4 \\
ATI & L1 (Reflex Only) & L1-L3  & L1-L4& L1-L3-L4 \\
ATI & L1/L2 (Full Stack) & L1/L2-L3  & L1/L2-L4 & L1/L2-L3-L4 \\
\bottomrule
\end{tabular}}
\label{tab:baselines}
\end{table}

We evaluate the prototype to answer three key questions:
(1) Can L2 effectively learn to stabilize inputs?
(2) Does active sensing improve local L3 performance compared with standard smartphone sensing baselines?
(3) Can ATI improve overall performance while reducing reliance on remote L4 inference?

\subsection{Model Configurations}

To isolate the contribution of each ATI component, we compare twelve configurations formed by combining four sensing strategies (AE, EIS, L1, and  L1/L2) with three inference paths (L3, L4, and L3-L4), as shown in Table~\ref{tab:baselines}.  
Here, AE denotes auto-exposure, and EIS denotes vendor-provided electronic image stabilization (i.e., Samsung's \textit{Super Steady} mode on Galaxy S25). The split inference setting (L3-L4) uses $\tau_{conf}=0.5$ and $\tau_{valid}=20$ for L4 escalation. 
This design disentangles the effects of (i) standard automatic capture vs.\ vendor stabilization (AE vs.\ EIS), (ii) replacing vendor stabilization with rule-based safety reflexes (EIS vs.\ L1), (iii) learned sensor stabilization (L1 vs.\ L1/L2), and (iv) different operating points on the local--remote inference continuum (L3, L4, and L3-L4).

\subsection{Results: L2 Learning Dynamics}
\label{sec:eval_l2}

\begin{figure}[t]
    \centering
    \includegraphics[width=1\linewidth]{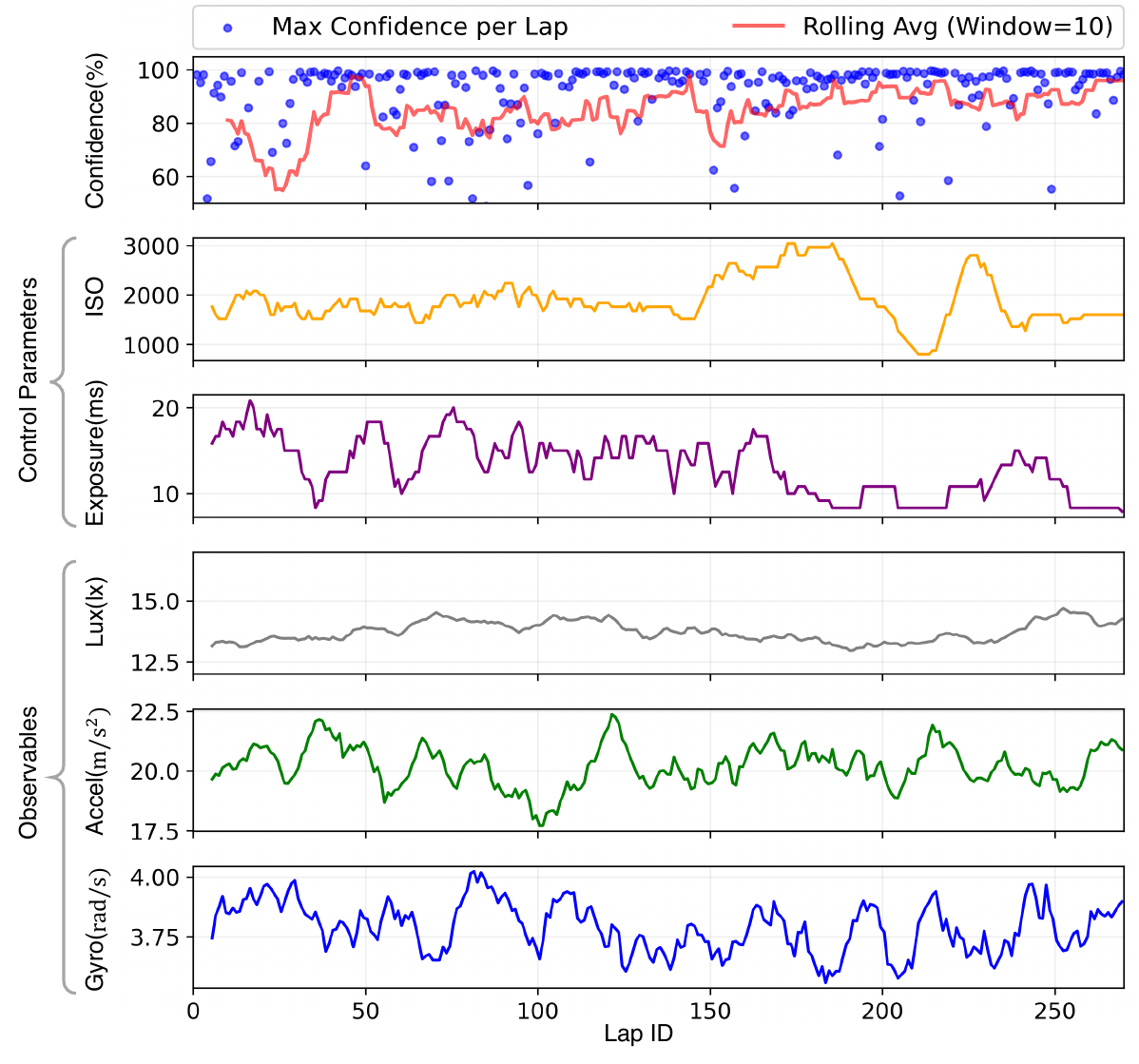}
    \caption{Evolution of camera parameters and confidence scores during L2 learning under low illumination. L2 adapts ISO and exposure within the safety envelope of L1. The rolling average of confidence score (red line) becomes higher and more stable over time, indicating  input stabilization.}
    \Description{Time-series plot showing ISO, exposure, and confidence during L2 learning under low illumination, with increasing confidence over time.}
    \label{fig:Cerebellum_learning}
\end{figure}

We conducted continuous learning experiments for a single object (i.e., Teddy) under two distinct illumination states: a bright environment ($\sim$150 lux) and sustained low illumination (below $\sim$15 lux). In this section, we focus on the low-light condition, which creates a challenging regime where motion blur and under-exposure must be balanced. Figure~\ref{fig:Cerebellum_learning} shows the evolution of camera parameters and the resulting confidence scores over 270 consecutive laps.

\noindent\textbf{Exploration to convergence.}
During the early phase (Laps 0--150), the bandit actively explores within the L1 safety envelope. In this stage, exposure frequently remains relatively long, and the per-lap confidence is highly variable, with repeated drops. Around Lap 160, the policy shifts toward shorter exposures, while ISO temporarily rises to compensate for reduced light intake. This transition is followed by a clear increase in the rolling confidence, which becomes more stable and remains mostly above 90\% in the later phase.

\begin{table*}[t]
    \centering
    \caption{Classification accuracy and L4 usage across twelve configurations formed by combining four sensing strategies (AE, EIS, L1, L1/L2) with three inference paths (L3, L4, and L3-L4). Per-class and total accuracies are reported over eight object classes. L3 uses EfficientNet-Lite0 as the local model, and L4 denotes the remote inference path evaluated via offline replay. The L4 call rate is measured on a per-lap basis and reports the percentage of laps processed by L4.}
    \renewcommand{\arraystretch}{1} 
    \setlength{\tabcolsep}{4pt}       
    \resizebox{\textwidth}{!}{%
        \begin{tabular}{llcccccccccc}
            \toprule
            \multirow{2}{*}{\textbf{Method}} & \multirow{2}{*}{\textbf{Model Config}} & \multicolumn{8}{c}{\textbf{Per-Class Accuracy (\%)}} & \multirow{2}{*}{\textbf{Total}} & \textbf{L4 Call} \\
            \cmidrule(lr){3-10}
             & & Teddy & Racket & Tennis Ball & Ping-pong Ball & Orange & Carton & Water Bottle & Laptop & & \textbf{Rate (\%)} \\
            \midrule
            \midrule
            \multirow{6}{*}{\textbf{Baselines}} 
             & AE-L3        & 84.0 & 32.0 & 90.0 & 94.0 & 16.0 & 28.0 & 34.0 & 22.0 & 50.0  & - \\
             & AE-L4        & 94.0 & 26.0 & 86.0 & 18.0 & 76.0 & 84.0 & 86.0 & 80.0 & 68.8 & 100.0 \\
             & AE-L3-L4     & 84.0 & 32.0 & 88.0 & 88.0 & 22.0 & 38.0 & 44.0 & 34.0 & 53.8 & 56.0 \\
             \cmidrule(lr){2-12}
             & EIS-L3       & 46.0 & 0.0  & 100.0& 0.0  & 0.0  & 2.0  & 8.0  & 0.0  & 20.0  & - \\
             & EIS-L4       & 94.0 & 4.0  & 86.0 & 0.0  & 76.0 & 10.0 & 88.0 & 46.0 & 51.0  & 100.0 \\
             & EIS-L3-L4    & 48.0 & 0.0  & 100.0& 2.0  & 4.0  & 4.0  & 8.0  & 0.0  & 21.0  & 85.0 \\
            \midrule                        
            \multirow{6}{*}{\textbf{ATI variants}} 
             & L1-L3        & 100.0 & 84.0 & 98.0 & 100.0 & 14.0 & 78.0 & 90.0 & 58.0 & 77.8 & - \\
             & L1-L4        & 90.0 & 54.0 & 94.0 & 36.0 & 90.0 & 94.0 & 92.0 & 92.0 & 80.3 & 100.0 \\
             & L1-L3-L4     & 100.0 & 80.0 & 100.0 & 98.0 & 38.0 & 76.0 & 90.0 & 70.0 & 81.5 & 32.8 \\
            \cmidrule(lr){2-12}
             & L1/L2-L3     & 100.0 & 94.0 & 100.0 & 92.0 & 26.0 & 84.0 & 84.0 & 62.0 & 80.3 & - \\
             & L1/L2-L4     & 88.0 & 74.0 & 96.0 & 36.0 & 92.0 & 90.0 & 86.0 & 94.0 & 82.0 & 100.0 \\
             & \textbf{L1/L2-L3-L4 (Ours)} & 100.0 & 92.0 & 100.0 & 94.0 & 62.0 & 96.0 & 92.0 & 68.0 & \textbf{88.0} & \textbf{31.8} \\
            \bottomrule
        \end{tabular}
    }
    \label{tab:model_configurations}
\end{table*}

\noindent\textbf{Steady-state behavior.}
After this transition, the policy settles into a more consistent operating regime. In the final phase (roughly Laps 250+), exposure remains short while ISO returns to a moderate range, indicating that L2 has learned a stable balance between blur suppression and sensor noise. 
Correspondingly, the rolling confidence stays near 95\% with substantially smaller fluctuations than in the initial phase. The remaining variation is mainly associated with changes in object visibility across laps rather than large shifts in sensing parameters.

\noindent\textbf{Dependence on the sensing regime.}
In contrast, repeating this learning experiment under bright illumination ($\sim$150 lux) yielded only marginal gains over the baseline (not shown). This indicates that L2 is most beneficial in physically constrained regimes, particularly under continuous motion and low light, where the trade-off between exposure time and gain becomes more critical.

\subsection{Results: Accuracy and Efficiency}

\begin{figure}[t]
    \centering
    \includegraphics[width=.85\columnwidth]{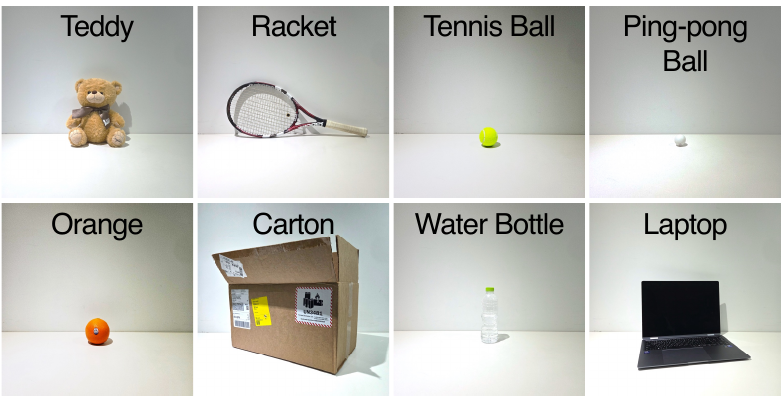} 
    \caption{Sample images of the eight object classes used in the experiment.}
    \Description{Sample images of the eight object classes used in the experiment.}
    \label{fig:object_classes}
\end{figure}

Table~\ref{tab:model_configurations} reports classification accuracy over the eight object classes shown in Figure~\ref{fig:object_classes}, together with the L4 usage rate for all twelve configurations, using EfficientNet-Lite0 as the default L3 model. 
Semantically equivalent output labels (e.g., \textit{notebook} and \textit{laptop}) were merged during evaluation.
The learned L2 internal model, trained on Teddy for 270 laps as in Section~\ref{sec:eval_l2}, was then reused across all eight objects for 50 laps each.

\noindent\textbf{Accuracy--efficiency trade-off across configurations.} 
Among all split/routing (L3-L4) configurations, L1/L2-L3-L4 performs best, reaching 88\% total accuracy with a 31.8\% L4 call rate. Compared with AE-L3-L4, it improves accuracy substantially while reducing reliance on remote inference. This indicates that better sensing increases the fraction of laps that can be resolved locally by L3.

\noindent\textbf{Contribution of L1 and L2.} 
The benefit of the sensing stack is already visible before routing is introduced. Replacing AE with L1 substantially improves L3-only performance (50\% to 77.8\%), showing that rule-based sensing constraints alone provide a large gain. Adding L2 further improves L3-only performance to 80.3\%, indicating that learned sensor adaptation contributes beyond fixed safety rules. The same pattern appears in the L3-L4 setting, where L1/L2 improves over L1 with nearly the same L4 call rate. Together, these results show that L1 and L2 play complementary roles: L1 avoids clearly harmful operating points, while L2 refines sensing within that safe region to improve downstream classification.

\noindent\textbf{Limitations of vendor stabilization (EIS).} 
The EIS baseline performs poorly in this setting, reaching only 21.0\% accuracy in the L3-L4 configuration while invoking L4 on 85.0\% of laps. In our prototype, EIS is implemented using Samsung's \textit{Super Steady} mode, which reduces apparent motion but also narrows the effective field of view and favors shorter exposures. Under low illumination, this leads to overly dark inputs that degrade L3 performance and trigger frequent escalation to L4. These results show that current vendor stabilization is not an effective substitute for explicit sensor control in low-light motion regimes.

\begin{table}[t]
\centering
\caption{Classification accuracy across baseline and ATI configurations with different local backbones. L4 call rates for hybrid L3-L4 settings are shown in parentheses.}
\label{tab:ati_backbone_comparison}
\renewcommand{\arraystretch}{1}
\resizebox{\columnwidth}{!}{%
\begin{tabular}{llcc}
\toprule
\textbf{Method} & \textbf{Model Config} & \textbf{MobileNet V2} & \textbf{MobileNet V3} \\
\midrule
\multirow{2}{*}[-0.5ex]{Baselines}
& AE-L3 & 42.8 & 28.5 \\
& AE-L3-L4 
& \begin{tabular}[c]{@{}c@{}}48.5 \\[-2pt] {\footnotesize (L4 call 58.8\%)}\end{tabular}
& \begin{tabular}[c]{@{}c@{}}34.8 \\[-2pt] {\footnotesize (L4 call 52.3\%)}\end{tabular} \\
\midrule
\multirow{4}{*}[-1.5ex]{ATI variants}
& L1-L3 & 58.8 & 51.0 \\
& L1-L3-L4 
& \begin{tabular}[c]{@{}c@{}}62.3 \\[-2pt] {\footnotesize (L4 call 33.8\%)}\end{tabular}
& \begin{tabular}[c]{@{}c@{}}52 \\[-2pt] {\footnotesize (L4 call 37.8\%)}\end{tabular} \\
\cmidrule{2-4}
& L1/L2-L3 & 66.0 & 53.3 \\
& \textbf{L1/L2-L3-L4 (Ours)}
& \begin{tabular}[c]{@{}c@{}}71.5 \\[-2pt] {\footnotesize (L4 call 29.0\%)}\end{tabular}
& \begin{tabular}[c]{@{}c@{}}64.8 \\[-2pt] {\footnotesize (L4 call 47.8\%)}\end{tabular} \\
\bottomrule
\end{tabular}%
}
\end{table}

\noindent\textbf{Why split inference outperforms L4-only inference.}
Within ATI, L1/L2-L3-L4 outperforms both L1/L2-L3 and L1/L2-L4, indicating that neither L3-only nor L4-only inference is sufficient by itself. The gain comes from selective routing and conditional L4 acceptance. 
This design is important because L4, implemented with Gemini, is a general-purpose remote model rather than a task-specialized classifier. As a result, even when prompted to return ImageNet labels, it can produce semantically plausible but task-misaligned predictions for visually similar objects (e.g., \textit{Ping-pong Ball} as ``golf ball''). By preserving confident L3 decisions while invoking and accepting L4 only when needed and more valuable, routing improves overall accuracy with minimal remote calls.

\noindent\textbf{Consistency across backbones.} 
Table~\ref{tab:ati_backbone_comparison} shows a similar trend for MobileNet V2 and MobileNet V3. In both cases, L1 improves over AE, and L2 provides an additional gain. The hybrid (L3-L4) ATI configuration also outperforms the corresponding AE hybrid baseline. These results suggest that the benefit of improved sensing is consistent across backbones, although the absolute gain depends on the strength of the local L3 model. Specifically, the lower accuracy of MobileNet V3 is due to the misclassification of ping-pong balls as pool table features, a limitation less evident in V2. A similar trend in L2 convergence was also observed across EfficientNet-Lite0, MobileNet V2, and MobileNet V3, suggesting that the learned L2 policy is not tied to a single backbone.

\subsection{Ablation: Threshold for L3 to L4 Escalation}
\label{sec:threshold_ablation}

\begin{figure}[t]
    \centering
    \includegraphics[width=1\linewidth]{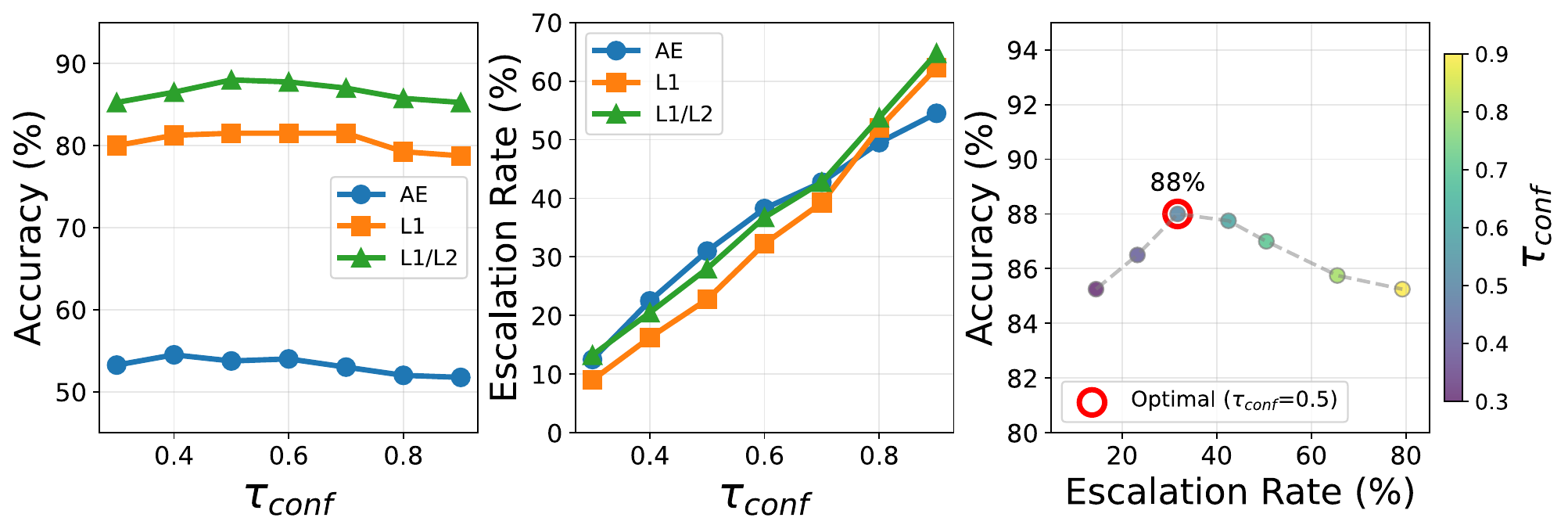}
    \caption{Ablation of the L3 escalation threshold $\tau_{conf}$ (EfficientNet-Lite0): (Left) Accuracy. (Middle) Escalation rate. (Right) Accuracy--escalation trade-off; best at $\tau_{conf}=0.5$.}
    \Description{Three plots showing accuracy, escalation rate, and their trade-off across L3 escalation thresholds, with the best result at $\tau_{conf}=0.5$.}
    \label{fig:th_ablation}
\end{figure}

We vary the L3 escalation threshold $\tau_{conf}$ from 0.3 to 0.9 with EfficientNet-Lite0 as the local model.
Figure~\ref{fig:th_ablation} shows that accuracy is highest around $\tau_{conf}=0.5$--$0.6$, whereas the escalation rate increases monotonically with $\tau_{conf}$.
The best trade-off is obtained at $\tau_{conf}=0.5$, which yields 88.0\% accuracy with a 31.8\% escalation rate.
We thus use $\tau_{conf}=0.5$ as the default threshold.

\subsection{Fast Sensor Adaptation Under Dynamic Lighting}

The previous experiments considered relatively stable lighting conditions, so state information such as illuminance was represented by the average measurement from the previous lap. 
To test whether ATI can also react to abrupt environmental changes, we alternated illumination between dark ($<15$ lux) and bright ($>150$ lux) conditions at the start of each lap, and compared ATI against the standard Auto Exposure (AE) baseline on the Teddy object. For ATI, we reused the converged L2 \textit{internal model} from Section~\ref{sec:eval_l2}, which triggers sensor-state updates immediately after each lighting transition, without waiting for L3's feedback.

\begin{figure}[t]
    \centering
    \includegraphics[width=1\linewidth]{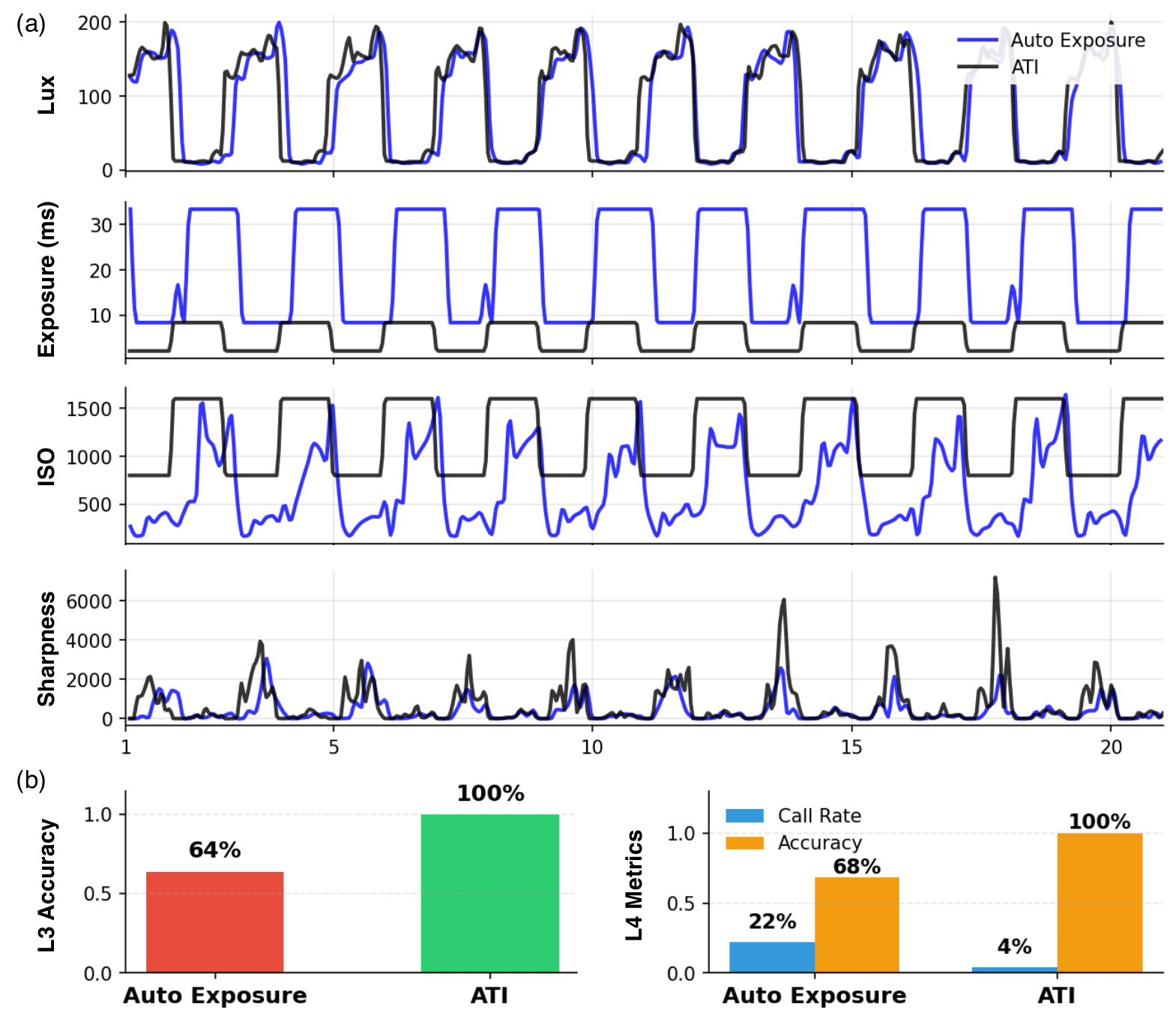}
    \caption{Performance under dynamic lighting on the Teddy object. (a) Time-series traces of illuminance, exposure time, ISO, and sharpness for Auto Exposure (AE, blue) and ATI (black) under alternating bright and dark conditions. (b) Comparison of L3 accuracy (left) and L4 call rate/final accuracy (right) between AE and ATI.}
    \Description{Comparison of Auto Exposure and ATI under dynamic lighting: time-series traces on the left and cascade performance metrics on the right.}
    \label{fig:dynamic_lighting_adapt}
\end{figure}

As shown in Figure~\ref{fig:dynamic_lighting_adapt}, both methods observe nearly identical lux transitions, but their camera responses differ substantially. Under low illumination, AE tends to increase exposure time aggressively to recover brightness, while its ISO fluctuates more sensitively across laps. In contrast, ATI maintains a shorter-exposure regime and adjusts ISO in a more coordinated manner, yielding inputs that are more suitable for downstream recognition.
This faster sensing loop directly improves downstream performance. ATI achieves 100\% accuracy at L3, compared with 64\% for AE, and reduces the L4 call rate from 22\% to 4\%. AE escalates more often, but the lower quality of the forwarded inputs limits the benefit of remote inference. These results show that ATI remains effective under rapid lighting changes that require fast on-device sensor control.

\subsection{Qualitative Analysis}

\begin{figure}[t]
    \centering
    \includegraphics[width=1\columnwidth]{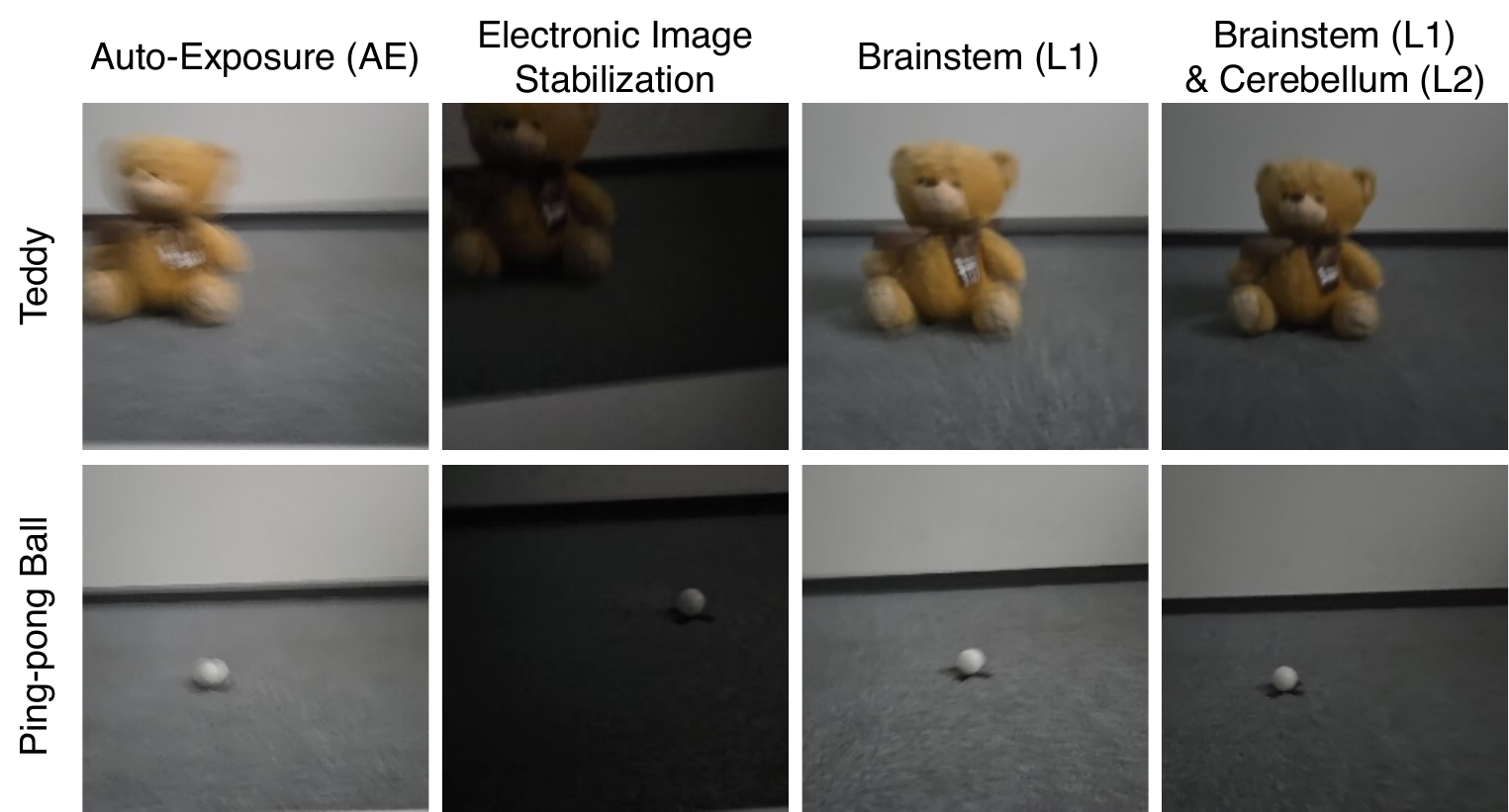} 
    \caption{Qualitative comparison of captured frames under different sensing strategies: Auto Exposure (AE), vendor electronic image stabilization (EIS), L1, and L1/L2.}
    \Description{Qualitative comparison of captured frames under four sensing strategies: Auto Exposure, vendor electronic image stabilization, L1, and L1/L2.}
    \label{fig:qualitative}
\end{figure}

Figure~\ref{fig:qualitative} compares representative frames captured under different sensing strategies.
Standard AE produces substantial motion blur under dynamic conditions.
EIS reduces apparent blur, but does so by narrowing the effective field of view through cropping and by darkening the image.
L1 improves over AE by enforcing motion-aware safety constraints, although residual blur remains under challenging conditions.
The full L1/L2 stack provides the most balanced result, preserving object visibility while better controlling both blur and exposure.
This qualitatively supports the quantitative results, suggesting that the learned active sensing policy produces inputs that are more suitable for the downstream task than the standard AE baseline.
\section{Discussion and Open Challenges} 
\label{sec:discussion}

ATI reframes physical AI not as a race to scale models, but as an architectural contract: preserve signal quality early, deliver the first useful decision locally, and invoke deeper reasoning only when the expected benefit justifies the added cost. Although this paper focuses on a vision prototype, the same principles of reflexive protection, predictive calibration, routine local handling, and selective escalation extend across sensing modalities. 
Here, we discuss how ATI generalizes, the open questions it raises, and the interfaces and benchmarks needed to make sensor-first intelligence practical.

\subsection{Generalization Beyond Vision}

The ATI decomposition applies whenever a sensor exposes meaningful control knobs and those choices materially affect downstream performance (Table~\ref{tab:sensing}). High-capacity models already exist for speech and audio~\cite{gong21b_interspeech, pham2019vdsn}, tactile sensing~\cite{chen2024tactile, fan2022graph}, and state estimation; what is missing is support below them for reflexive protection and continuous calibration.

\begin{figure}[t]
    \centering
    \includegraphics[width=1\columnwidth]{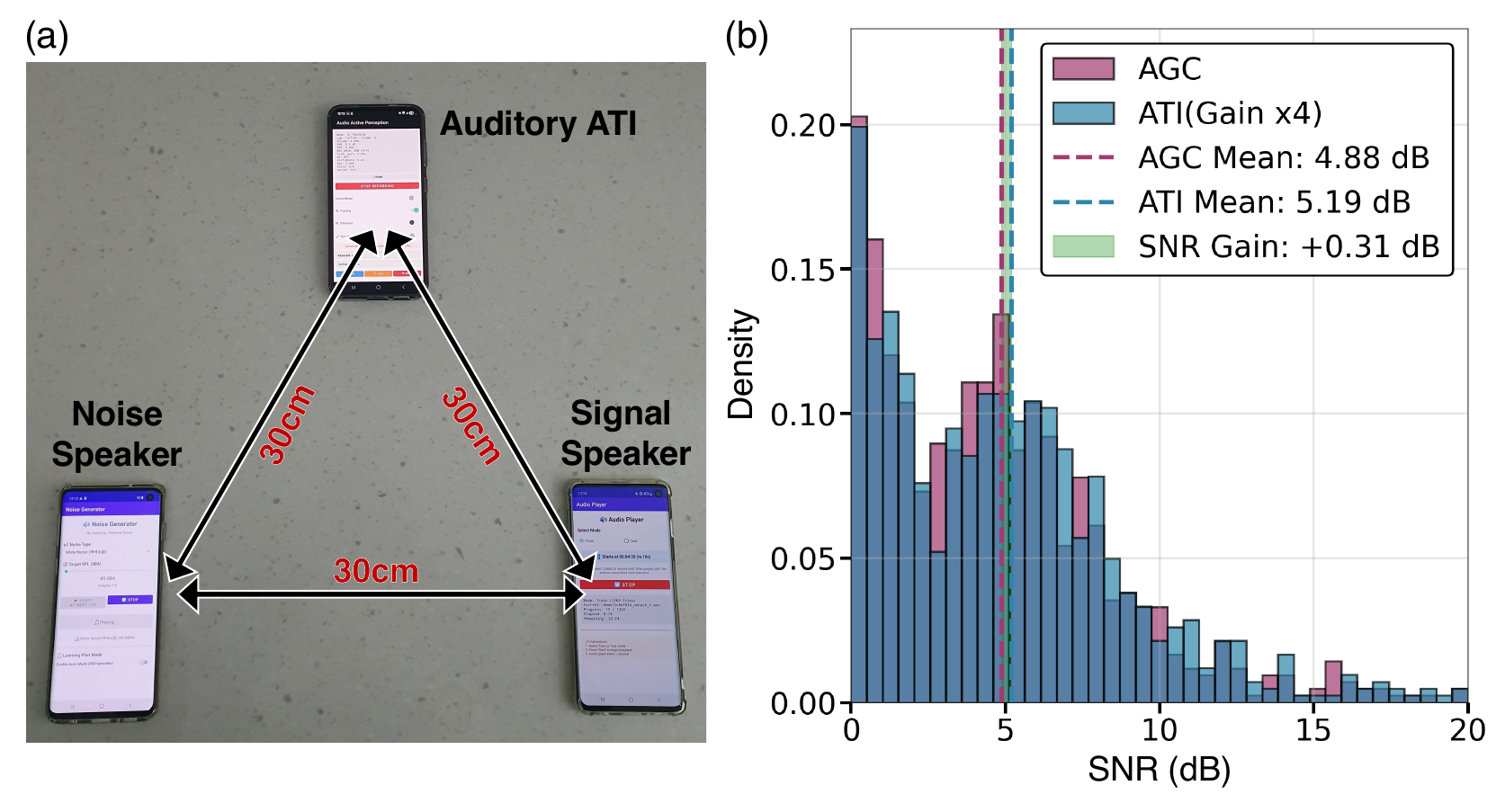} 
    \caption{Auditory ATI proof of concept. (a) Physical audio setup with one microphone, one signal speaker, and one noise speaker. (b) SNR distributions for AGC and ATI under a noise-dominant condition.}
    \Description{Auditory ATI proof of concept: physical audio setup on the left and SNR distributions for AGC and ATI on the right.}
    \label{fig:audio}
\end{figure}

\noindent \textbf{Auditory ATI (a ``smart cochlea'').} 
Conventional audio pipelines typically apply gain control and denoising \textit{after capture}. ATI instead moves part of this adaptation into the sensing front end.
\begin{itemize}[leftmargin=*, nosep] 
    \item \textbf{L1:} L1 provides short-timescale clipping protection, analogous to the stapedius reflex~\cite{moller2012hearing}, by adjusting analog gain to prevent saturation during transient spikes.

    \item \textbf{L2:} L2 performs predictive audio calibration by tuning gain, filtering, and beamforming in real time. Analogous to the function of outer hair cells~\cite{ashmore2008cochlear}, it amplifies task-relevant signals and suppresses nuisance noise before heavy DSP or speech models are invoked.
\end{itemize}
\noindent
As a preliminary proof of concept, a CMAB-based L2 calibration policy adjusted digital gain and noise suppression for a downstream keyword recognizer on Google Speech Commands v0.01.
In the physical setup shown in Figure~\ref{fig:audio}(a), with one microphone, one signal speaker, and one noise speaker, Auditory ATI achieved a mean SNR gain of +0.31 dB over a passive Automatic Gain Control (AGC) baseline, as shown in Figure~\ref{fig:audio}(b). This improvement in signal quality increased keyword recognition accuracy from 31.6\% to 32.7\%. Although modest, this result supports ATI's main claim that learned front-end calibration can improve downstream perception.

\noindent \textbf{Tactile ATI (active touch).} 
Tactile sensing is inherently interactive: resolving texture, compliance, or shape often requires motion and exploratory contact~\cite{lederman2009haptic, johansson2009coding}.
\begin{itemize}[leftmargin=*, nosep] 

    \item \textbf{L1} would monitor contact pressure at high frequency and trigger reflex withdrawal under sharp impact, analogous to nociceptive withdrawal reflexes~\cite{skljarevski2002nociceptive}. This would protect the tactile sensor and end effector before a higher-level planner can react.
    
    \item \textbf{L2} would manage active touch, for example by selecting regions of interest on the tactile array, modulating contact force, or adjusting exploratory motion to increase signal informativeness.
\end{itemize}

\noindent \textbf{Proprioceptive ATI (a ``smart IMU'').} 
IMUs are often treated as passive sensors followed by fixed filtering. ATI suggests a more adaptive alternative.
\begin{itemize}[leftmargin=*, nosep] 

    \item \textbf{L1} would enforce kinematic safety limits, e.g., emergency stop when acceleration or joint velocity exceeds hardware bounds.
    
    \item \textbf{L2} would adapt filter parameters and estimate bias online, suppressing vibration during motion while increasing sensitivity when the system is stationary, analogous to the adaptability of the biological vestibular system.
\end{itemize}

\noindent
More generally, ATI applies whenever a sensor \textit{exposes meaningful control knobs}, such as gain, integration time, beam pattern, contact force, taxel selection, or filter bandwidth, and those choices materially affect downstream performance.

\subsection{Open Algorithmic Questions for Sensor-First Intelligence}

Our current prototype studies a minimal ATI setting with one sensor, one L3 routine (classification), and lap-level escalation to L4. In a more general ATI system, coupling sensing, routine local handling, and deep reasoning raises several new algorithmic questions.

\noindent\textbf{Reliable proxies for L3 uncertainty.} 
ATI currently uses raw softmax confidence as a routing signal, which is often overconfident. Better proxies are needed for the question, ``Is L3 truly reliable on this input?'' and those proxies will likely be task-specific (e.g., detection, segmentation, pose estimation, and visual question answering (VQA)), combining semantic confidence with cues such as localization quality, temporal stability, or cross-modal consistency.

\noindent\textbf{Controlling one sensor for multiple downstream models.} 
In real systems, one camera may simultaneously feed classification, detection, tracking, VQA, and control. L2 must then optimize a shared sensor for multiple consumers with different latency budgets and notions of utility, raising a multi-objective control problem.

\noindent\textbf{Coupling L2 with test-time adaptation in L3.} 
Once L3 can adapt online, L2 may need to shape an input distribution that allows L3 to adapt quickly and stably, rather than simply maximizing the current prediction. This creates a co-adaptation problem between sensor control and online model updates.

\noindent\textbf{Latency-constrained L2 control.} 
Our dynamic-lighting experiment already shows that fast sensor control matters. More generally, camera actions such as shutter time, burst capture, ROI readout, autofocus, and active illumination affect both signal quality and when the next usable observation becomes available. L2 must therefore optimize sensing under deadline while accounting for sensing delay, downstream latency, and task urgency.

\noindent\textbf{Offloading under live network variation.} 
A general ATI system must decide whether to resample, wait for L2 to stabilize the signal, or escalate to L4 under fluctuating network conditions and under L2 actions that themselves consume time. What to offload also becomes less obvious when temporal differences may reflect sensor adaptation rather than scene change.

\noindent\textbf{Compression for sensor-adaptive local models.} 
If L2 can steer inputs toward model-friendly regimes, the best compressed L3 model may differ from the one chosen for a passive pipeline. This suggests a new co-design problem between model compression and sensor control.

\subsection{Systems Substrate, Metrics, and Benchmarks}

For ATI to mature from an architectural idea into a portable eco-system, the systems community needs common interfaces, evaluation protocols, and shared data.

\noindent \textbf{Standardizing ATI interfaces.} 
Three interfaces appear especially important: (1) compact quality-vector schemas through which L2 communicates signal health to L3; (2) advisory escalation protocols with explicit time-to-live semantics so cloud responses can refine local decisions without blocking safety-critical loops; and (3) machine-checkable safety-envelope specifications for L1 so learned policies remain bounded by hardware limits.

\noindent\textbf{Metrics beyond static accuracy.} 
Standard metrics such as classification accuracy and mAP do not capture the dynamics of sensor-first systems. ATI-style systems should also report time-to-first-decision (TTFD), deadline miss rate, prevention rate, escalation utility, adaptation stability, energy per useful observation, and sensor-induced latency. ATI should reward not only correct answers, but also correct and timely sensing behavior.

\noindent\textbf{Benchmarks and tooling.} 
ATI needs benchmarks that vary the sensor as well as the scene. For visual ATI, this means sequences captured under controlled sweeps of exposure, ISO, focus, white balance, motion, illumination, and network conditions, ideally with task deadlines representative of robotics, assistive vision, and smart-glass workloads. More generally, ATI benchmarks should support multiple downstream tasks, including classification~\cite{baek2024unexplored,baekadaptive}, detection, segmentation, pose estimation~\cite{han2025senseshift6d,yoon2026ego}, and language-grounded queries, so that ATI can be studied as a multi-task architecture rather than a single-model pipeline.
Progress will also require simulators that capture sensor physics, noise, blur, saturation, latency, and control delay well enough to train and compare L2 policies offline, ideally coupled with reliable sim2real pipelines for sensor variation~\cite{kim2026imagenet} and trace-driven network and systems simulators.

\subsection{Risks and Limitations}

\noindent\textbf{Adversarial or unstable sensor control.} 
Granting learned agents control over hardware introduces new failure modes. Adversarial physical patterns, such as stroboscopic lights or carefully timed audio bursts, could drive L2 into oscillation, self-blinding, or other pathological behaviors. This is why L1 must be treated as part of the trusted computing base: its safety envelopes must remain non-bypassable regardless of the learned policy or the input.

\noindent\textbf{Scope of applicability.} 
ATI is not a universal replacement for monolithic models. It is most valuable when the sensor exposes meaningful control knobs, the environment is dynamic, and task difficulty is skewed so that most cases can be handled locally while a smaller fraction require deeper reasoning. In static environments with abundant power, compute, and stable sensing, a single monolithic model may be simpler and entirely adequate. ATI is aimed specifically at the physical regime, where early control over data acquisition materially changes system behavior.
\section{Conclusion}

Physical AI exposes a limitation of today's computation-centric AI stack: when sensing is fixed, downstream models must compensate for failures that should have been prevented at capture time. 
We proposed Artificial Tripartite Intelligence (ATI), a sensor-first architecture that organizes physical AI into Brainstem (L1) for reflexive safety and signal integrity, Cerebellum (L2) for continuous sensor calibration, and a Cerebral inference subsystem spanning L3/L4 for routine local skill selection and execution, coordination, and deeper reasoning. 
ATI closes the loop between sensing and inference under explicit latency, energy, and reliability constraints. 
More fundamentally, ATI reframes sensing not as a passive front-end to perception, but as an actively managed part of the intelligence stack.

Our commodity-smartphone prototype shows that this decomposition is practical: safety envelopes can bound sensor control, lightweight calibration can improve capture quality, and selective escalation can reserve expensive reasoning for cases that truly benefit from it. 
While demonstrated here on a mobile vision prototype, ATI is intended as a general sensor-first architectural framework for physical AI. 
More broadly, ATI points to a systems agenda: co-design sensing, local inference, and offloading, and build the interfaces, metrics, and benchmarks needed for sensor-first intelligence.

\begin{acks}
This work was supported in part by the Institute of Information \& communications Technology Planning \& Evaluation (IITP) grant funded by the Korea government (MSIT) (No. RS-2026-25507282, No. RS-2025-25463302), and in part by the National Research Foundation (NRF) of Korea grant funded by the Korea government (MSIT) (No. RS-2023-00212780, No. RS-2023-00222663, No. RS-2025-25407378). 
\end{acks}





\appendix
\section{Artifact Appendix}

\subsection{Abstract}

This artifact provides the device-side implementation of Artificial Tripartite Intelligence (ATI), a bio-inspired, sensor-first architecture for physical AI. 
ATI decomposes the intelligence stack into Brainstem (L1, reflex control), Cerebellum (L2, sensor calibration), and a cerebral inference subsystem spanning the Basal Ganglia Network (BGN, L3) for routine on-device skill selection and execution and the Hippocampal-Cortical Network (HCN, L4) for deeper reasoning. 
This design preserves functional separation between fast sensor-side adaptation (L1/L2) and higher-level inference (L3/L4), while coordinating them through explicit interfaces.
This repository contains the Android application (Kotlin) that implements millisecond-scale L1 reflexes, sensor-rate L2 calibration, and routine L3 execution using on-device TFLite models, together with the logging and control pipeline used in the ATI prototype.

\subsection{Artifact check-list (meta-information)}

{\small
\begin{itemize}
  \item {\bf Algorithm: } Contextual Multi-Armed Bandit (CMAB) for sensor calibration.
  \item {\bf Program: } Android Application (Kotlin).
  \item {\bf Compilation: } Android Studio Hedgehog (2023.1.1+) and Gradle 8.2+.
  \item {\bf Binary: } Android APK.
  \item {\bf Model: } TensorFlow Lite (MobileNet V2, MobileNet V3, EfficientNet-Lite0).
  \item {\bf Data set: } Real-time sensor logs and prediction confidence generated during runtime.
  \item {\bf Run-time environment: } Android 12 (API 31) or higher.
  \item {\bf Hardware: } Android smartphone (e.g., Samsung Galaxy S25) with Camera2 API.
  \item {\bf Execution: } Standalone app execution with manual UI interaction.
  \item {\bf Metrics: } Classification accuracy, RL reward convergence, and performance profiling.
  \item {\bf Output: } Log files (AP\_log.csv, RL\_log.csv), learned policy table, and compressed PNG images.
  \item {\bf Experiments: } Navigation laps under varying illumination conditions (bright/dark).
  \item {\bf How much disk space required (approximately)?: } 1GB free space.
  \item {\bf How much time is needed to prepare workflow (approximately)?: } 15-30 minutes for Android Studio setup and APK installation.
  \item {\bf How much time is needed to complete experiments (approximately)?: } 6 hours
  \item {\bf Publicly available?: } Yes.
  \item {\bf Code licenses (if publicly available)?: } Apache License 2.0.
  \item {\bf Archived (provide DOI)?: } 10.5281/zenodo.19145654.
\end{itemize}
}

\subsection{Description}

\subsubsection{How to access}

The source code is publicly accessible via GitHub (\url{https://github.com/sunake7/ATI}) and is permanently archived on Zenodo (DOI: 10.5281/zenodo.19145654).

\subsubsection{Hardware dependencies}

The application requires an Android device supporting the Camera2 API for manual control of ISO and shutter speed. For full physical navigation experiments, the device should be mounted on a mobile platform (e.g., a toy car) navigating a track under controllable lighting.

\subsubsection{Software dependencies}

The project requires Android Studio (Hedgehog 2023.1.1 or later), Gradle 8.2+, Kotlin 1.9+, Target SDK 36, and Min SDK 31.

\subsubsection{Data sets}

No external datasets are required to build or run the application. The system generates its own data logs (CSV format) during runtime.

\subsubsection{Models}

Pre-trained TFLite models (MobileNet V3 Large, MobileNet V2 1.0, EfficientNet-Lite0) for routine skill execution (L3) are included directly within the repository to support on-device inference.

\subsection{Installation}

\begin{enumerate}
\item Clone the repository from GitHub: \url{https://github.com/sunake7/ATI.git}
\item Open the project folder in Android Studio.
\item Allow Android Studio to automatically sync Gradle dependencies.
\item Build the APK using the \texttt{./gradlew assembleDebug} command or via the Android Studio Build menu.
\item Install the compiled APK onto a target Android device using ADB (\texttt{adb install app/build/outputs/apk/debug/\allowbreak app-debug.apk}).
\end{enumerate}

\subsection{Experiment workflow}

The general workflow involves launching the app, granting camera permissions, selecting the desired local AI model, configuring the intelligence mode using the UI toggles, and tapping the START LOG button. The app then captures frames, applies the designated sensor control policy, runs local inference, and saves the generated data to the device storage.

\subsection{Evaluation and expected results}

Upon successful compilation and execution, users can navigate the app UI to configure the intelligence modes using toggle buttons: Active Base (Brainstem), RL Agent (Cerebellum) for RL training, RL Agent (infer) for trained RL inference only, and Test mode. Disabling the Active Base and RL Agent toggles defaults the system to Passive (Auto Exposure) mode. If Test mode is initiated, the application automatically executes a sequential inference session, running 50 laps each in Passive, Active Base, and RL Agent (infer) modes. The application generates detailed CSV files in the device's Downloads folder documenting real-time sensor states and classification confidence scores. Crucially, RL-specific logs, such as agent actions and rewards, are exclusively saved when the RL Agent (Cerebellum) training mode is selected. Additionally, the 224x224 images used during local inference are captured and saved as PNG files. When the experiment concludes, these images are compressed into an archive and saved alongside the CSV log files. These exported PNG images are intended to be used for remote L4 (Hippocampal-Cortical Network) deep reasoning inference.


\bibliographystyle{ACM-Reference-Format}
\bibliography{reference}

@String{Computing = "Computing" }

@String{Computer = "{IEEE} Computer" }

@article{brown2020language,
  title={Language models are few-shot learners},
  author={Brown, Tom and Mann, Benjamin and Ryder, Nick and Subbiah, Melanie and Kaplan, Jared D and Dhariwal, Prafulla and Neelakantan, Arvind and Shyam, Pranav and Sastry, Girish and Askell, Amanda and others},
  journal={Advances in neural information processing systems},
  volume={33},
  pages={1877--1901},
  year={2020}
}

@article{vaswani2017attention,
  title={Attention is all you need},
  author={Vaswani, Ashish and Shazeer, Noam and Parmar, Niki and Uszkoreit, Jakob and Jones, Llion and Gomez, Aidan N and Kaiser, {\L}ukasz and Polosukhin, Illia},
  journal={Advances in neural information processing systems},
  volume={30},
  year={2017}
}

@inproceedings{radford2021learning,
  title={Learning transferable visual models from natural language supervision},
  author={Radford, Alec and Kim, Jong Wook and Hallacy, Chris and Ramesh, Aditya and Goh, Gabriel and Agarwal, Sandhini and Sastry, Girish and Askell, Amanda and Mishkin, Pamela and Clark, Jack and others},
  booktitle={International conference on machine learning},
  pages={8748--8763},
  year={2021},
  organization={PmLR}
}

@inproceedings{park2023pointsplit,
  title={PointSplit: Towards on-device 3D object detection with heterogeneous low-power accelerators},
  author={Park, Keondo and Choi, You Rim and Lee, Inhoe and Kim, Hyung-Sin},
  booktitle={Proceedings of the 22nd International Conference on Information Processing in Sensor Networks},
  pages={67--81},
  year={2023}
}

@article{im2025tokens,
  title={From Tokens to Photons: Test-Time Physical Prompting for Vison-Language Models},
  author={Im, Boyeong and Lee, Wooseok and Kwon, Yoojin and Kim, Hyung-Sin},
  journal={arXiv preprint arXiv:2512.12571},
  year={2025}
}

@inproceedings{choi2022scriptpainter,
  title={ScriptPainter: Vision-based, on-device test script generation for mobile systems},
  author={Choi, Yousung and Seo, Ahreum and Kim, Hyung-Sin},
  booktitle={2022 21st ACM/IEEE International Conference on Information Processing in Sensor Networks (IPSN)},
  pages={477--490},
  year={2022},
  organization={IEEE}
}

@article{kang2024mirror,
  title={Mirror: Towards generalizable on-device video virtual try-on for mobile shopping},
  author={Kang, Dong-Sig and Baek, Eunsu and Son, Sungwook and Lee, Youngki and Gong, Taesik and Kim, Hyung-Sin},
  journal={Proceedings of the ACM on Interactive, Mobile, Wearable and Ubiquitous Technologies},
  volume={7},
  number={4},
  pages={1--27},
  year={2024},
  publisher={ACM New York, NY, USA}
}

@article{park2025distillsleep,
  title={DistillSleep: real-time, on-device, interpretable sleep staging from single-channel electroencephalogram},
  author={Park, Keondo and Hong, Joopyo and Lee, Wooseok and Shin, Hyun-Woo and Kim, Hyung-Sin},
  journal={SLEEPJ},
  volume={48},
  number={12},
  pages={zsaf240},
  year={2025},
  publisher={Oxford University Press}
}

@inproceedings{youn2022bitwidth,
  title={Bitwidth-adaptive quantization-aware neural network training: A meta-learning approach},
  author={Youn, Jiseok and Song, Jaehun and Kim, Hyung-Sin and Bahk, Saewoong},
  booktitle={European Conference on Computer Vision},
  pages={208--224},
  year={2022},
  organization={Springer}
}

@inproceedings{chen2018marvel,
  title={Marvel: Enabling mobile augmented reality with low energy and low latency},
  author={Chen, Kaifei and Li, Tong and Kim, Hyung-Sin and Culler, David E and Katz, Randy H},
  booktitle={Proceedings of the 16th ACM Conference on Embedded Networked Sensor Systems},
  pages={292--304},
  year={2018}
}

@article{hu2023toward,
  title={Toward general-purpose robots via foundation models: A survey and meta-analysis},
  author={Hu, Yafei and Xie, Quanting and Jain, Vidhi and Francis, Jonathan and Patrikar, Jay and Keetha, Nikhil and Kim, Seungchan and Xie, Yaqi and Zhang, Tianyi and Fang, Hao-Shu and others},
  journal={arXiv preprint arXiv:2312.08782},
  year={2023}
}

@article{ma2024survey,
  title={A survey on vision-language-action models for embodied ai},
  author={Ma, Yueen and Song, Zixing and Zhuang, Yuzheng and Hao, Jianye and King, Irwin},
  journal={arXiv preprint arXiv:2405.14093},
  year={2024}
}

@article{liu2025aligning,
  title={Aligning cyber space with physical world: A comprehensive survey on embodied ai},
  author={Liu, Yang and Chen, Weixing and Bai, Yongjie and Liang, Xiaodan and Li, Guanbin and Gao, Wen and Lin, Liang},
  journal={IEEE/ASME Transactions on Mechatronics},
  year={2025},
  publisher={IEEE}
}

@article{zhang2024efficient,
  title={Efficient Camera Exposure Control for Visual Odometry via Deep Reinforcement Learning},
  author={Zhang, Shuyang and He, Jinhao and Zhu, Yilong and Wu, Jin and Yuan, Jie},
  journal={IEEE Robotics and Automation Letters},
  year={2024},
  publisher={IEEE}
}

@inproceedings{baekadaptive,
  title={Adaptive Camera Sensor for Vision Models},
  author={Baek, Eunsu and Han, Sung-hwan and Gong, Taesik and Kim, Hyung-Sin},
  booktitle={The Thirteenth International Conference on Learning Representations},
  year={2025}
}

@article{ito2008control,
  title={Control of mental activities by internal models in the cerebellum},
  author={Ito, Masao},
  journal={Nature Reviews Neuroscience},
  volume={9},
  number={4},
  pages={304--313},
  year={2008},
  publisher={Nature Publishing Group UK London}
}

@inproceedings{quigley2009ros,
  title={ROS: an open-source Robot Operating System},
  author={Quigley, Morgan and Conley, Ken and Gerkey, Brian and Faust, Josh and Foote, Tully and Leibs, Jeremy and Wheeler, Rob and Ng, Andrew Y and others},
  booktitle={ICRA workshop on open source software},
  volume={3},
  number={3.2},
  pages={5},
  year={2009},
  organization={Kobe}
}

@article{krizhevsky2012imagenet,
  title={Imagenet classification with deep convolutional neural networks},
  author={Krizhevsky, Alex and Sutskever, Ilya and Hinton, Geoffrey E},
  journal={Advances in neural information processing systems},
  volume={25},
  year={2012}
}

@article{thorpe1996speed,
  title={Speed of processing in the human visual system},
  author={Thorpe, Simon and Fize, Denis and Marlot, Catherine},
  journal={nature},
  volume={381},
  number={6582},
  pages={520--522},
  year={1996},
  publisher={Nature Publishing Group UK London}
}

@article{lamme2000distinct,
  title={The distinct modes of vision offered by feedforward and recurrent processing},
  author={Lamme, Victor AF and Roelfsema, Pieter R},
  journal={Trends in neurosciences},
  volume={23},
  number={11},
  pages={571--579},
  year={2000},
  publisher={Elsevier}
}

@article{ito1998cerebellar,
  title={Cerebellar learning in the vestibulo--ocular reflex},
  author={Ito, Masao},
  journal={Trends in cognitive sciences},
  volume={2},
  number={9},
  pages={313--321},
  year={1998},
  publisher={Elsevier}
}

@article{do2019melanopsin,
  title={Melanopsin and the intrinsically photosensitive retinal ganglion cells: biophysics to behavior},
  author={Do, Michael Tri H},
  journal={Neuron},
  volume={104},
  number={2},
  pages={205--226},
  year={2019},
  publisher={Elsevier}
}

@inproceedings{ichnowski2023fogros2,
  title={FogROS2: An Adaptive Platform for Cloud and Fog Robotics Using ROS 2},
  author={Ichnowski, Jeffrey and Chen, Kaiyuan and Dharmarajan, Karthik and Adebola, Simeon and Danielczuk, Michael and Mayoral-Vilches, V{\'\i}ctor and Jha, Nikhil and Zhan, Hugo and Llontop, Edith and Xu, Derek and others},
  booktitle={2023 IEEE International Conference on Robotics and Automation (ICRA)},
  pages={5493--5500},
  year={2023},
  organization={IEEE}
}

@article{bjorck2025gr00t,
  title={Gr00t n1: An open foundation model for generalist humanoid robots},
  author={Bjorck, Johan and Casta{\~n}eda, Fernando and Cherniadev, Nikita and Da, Xingye and Ding, Runyu and Fan, Linxi and Fang, Yu and Fox, Dieter and Hu, Fengyuan and Huang, Spencer and others},
  journal={arXiv preprint arXiv:2503.14734},
  year={2025}
}

@inproceedings{miluzzo2008sensing,
  title={Sensing meets mobile social networks: the design, implementation and evaluation of the cenceme application},
  author={Miluzzo, Emiliano and Lane, Nicholas D and Fodor, Krist{\'o}f and Peterson, Ronald and Lu, Hong and Musolesi, Mirco and Eisenman, Shane B and Zheng, Xiao and Campbell, Andrew T},
  booktitle={Proceedings of the 6th ACM conference on Embedded network sensor systems},
  pages={337--350},
  year={2008}
}

@inproceedings{lu2009soundsense,
  title={Soundsense: scalable sound sensing for people-centric applications on mobile phones},
  author={Lu, Hong and Pan, Wei and Lane, Nicholas D and Choudhury, Tanzeem and Campbell, Andrew T},
  booktitle={Proceedings of the 7th international conference on Mobile systems, applications, and services},
  pages={165--178},
  year={2009}
}

@inproceedings{cuervo2010maui,
  title={Maui: making smartphones last longer with code offload},
  author={Cuervo, Eduardo and Balasubramanian, Aruna and Cho, Dae-ki and Wolman, Alec and Saroiu, Stefan and Chandra, Ranveer and Bahl, Paramvir},
  booktitle={Proceedings of the 8th international conference on Mobile systems, applications, and services},
  pages={49--62},
  year={2010}
}

@article{satyanarayanan2009case,
  title={The case for vm-based cloudlets in mobile computing},
  author={Satyanarayanan, Mahadev and Bahl, Paramvir and Caceres, Ram{\'o}n and Davies, Nigel},
  journal={IEEE pervasive Computing},
  volume={8},
  number={4},
  pages={14--23},
  year={2009},
  publisher={IEEE}
}

@inproceedings{han2016mcdnn,
  title={Mcdnn: An approximation-based execution framework for deep stream processing under resource constraints},
  author={Han, Seungyeop and Shen, Haichen and Philipose, Matthai and Agarwal, Sharad and Wolman, Alec and Krishnamurthy, Arvind},
  booktitle={Proceedings of the 14th Annual International Conference on Mobile Systems, Applications, and Services},
  pages={123--136},
  year={2016}
}

@inproceedings{baekposition,
  title={Position: AI Should Sense Better, Not Just Scale Bigger: Adaptive Sensing as a Paradigm Shift},
  author={Baek, Eunsu and Park, Keondo and Ko, Jeonggil and Oh, Min-hwan and Gong, Taesik and Kim, Hyung-Sin},
  booktitle={The Thirty-Ninth Annual Conference on Neural Information Processing Systems Position Paper Track},
  year={2025}
}

@inproceedings{fang2018nestdnn,
  title={Nestdnn: Resource-aware multi-tenant on-device deep learning for continuous mobile vision},
  author={Fang, Biyi and Zeng, Xiao and Zhang, Mi},
  booktitle={Proceedings of the 24th Annual International Conference on Mobile Computing and Networking},
  pages={115--127},
  year={2018}
}

@article{macenski2022robot,
  title={Robot operating system 2: Design, architecture, and uses in the wild},
  author={Macenski, Steven and Foote, Tully and Gerkey, Brian and Lalancette, Chris and Woodall, William},
  journal={Science robotics},
  volume={7},
  number={66},
  pages={eabm6074},
  year={2022},
  publisher={American Association for the Advancement of Science}
}

@inproceedings{chen2024fogros2,
  title={FogROS2-FT: Fault Tolerant Cloud Robotics},
  author={Chen, Kaiyuan and Hari, Kush and Chung, Trinity and Wang, Michael and Tian, Nan and Juette, Christian and Ichnowski, Jeffrey and Ren, Liu and Kubiatowicz, John and Stoica, Ion and others},
  booktitle={2024 IEEE/RSJ International Conference on Intelligent Robots and Systems (IROS)},
  pages={1390--1397},
  year={2024},
  organization={IEEE}
}

@inproceedings{zitkovich2023rt,
  title={Rt-2: Vision-language-action models transfer web knowledge to robotic control},
  author={Zitkovich, Brianna and Yu, Tianhe and Xu, Sichun and Xu, Peng and Xiao, Ted and Xia, Fei and Wu, Jialin and Wohlhart, Paul and Welker, Stefan and Wahid, Ayzaan and others},
  booktitle={Conference on Robot Learning},
  pages={2165--2183},
  year={2023},
  organization={PMLR}
}

@inproceedings{jia2024empowering,
  title={Empowering in-browser deep learning inference on edge through just-in-time kernel optimization},
  author={Jia, Fucheng and Jiang, Shiqi and Cao, Ting and Cui, Wei and Xia, Tianrui and Cao, Xu and Li, Yuanchun and Wang, Qipeng and Zhang, Deyu and Ren, Ju and others},
  booktitle={Proceedings of the 22nd Annual International Conference on Mobile Systems, Applications and Services},
  pages={438--450},
  year={2024}
}

@article{team2025gemini,
  title={Gemini robotics: Bringing ai into the physical world},
  author={Team, Gemini Robotics and Abeyruwan, Saminda and Ainslie, Joshua and Alayrac, Jean-Baptiste and Arenas, Montserrat Gonzalez and Armstrong, Travis and Balakrishna, Ashwin and Baruch, Robert and Bauza, Maria and Blokzijl, Michiel and others},
  journal={arXiv preprint arXiv:2503.20020},
  year={2025}
}

@inproceedings{kim2026imagenet,
  title={ImageNet-sES: A First Systematic Study of Sensor-Environment Simulation Anchored by Real Recaptures},
  author={Kim, Ji-yoon and Baek, Eunsu and Kim, Hyung-Sin},
  booktitle={Proceedings of the IEEE/CVF Winter Conference on Applications of Computer Vision},
  pages={1117--1126},
  year={2026}
}

@inproceedings{baek2024unexplored,
  title={Unexplored faces of robustness and out-of-distribution: Covariate shifts in environment and sensor domains},
  author={Baek, Eunsu and Park, Keondo and Kim, Jiyoon and Kim, Hyung-Sin},
  booktitle={Proceedings of the IEEE/CVF Conference on Computer Vision and Pattern Recognition},
  pages={22294--22303},
  year={2024}
}

@inproceedings{yoon2026ego,
  title={EgoXtreme: A Dataset for Robust Object Pose Estimation  in Egocentric Views under Extreme Conditions},
  author={Taegyoon Yoon and Yegyu Han and Seojin Ji and Jaewoo Park and Sojeong Kim and Taein Kwon and Hyung-Sin Kim},
  booktitle={Proceedings of the IEEE/CVF Conference on Computer Vision and Pattern Recognition},
  year={2026}
}

@article{han2025senseshift6d,
  title={Senseshift6d: Multimodal rgb-d benchmarking for robust 6d pose estimation across environment and sensor variations},
  author={Han, Yegyu and Yoon, Taegyoon and Woo, Dayeon and Kim, Sojeong and Kim, Hyung-Sin},
  journal={arXiv preprint arXiv:2507.05751},
  year={2025}
}

@inproceedings{driess2023palm,
  title={PaLM-E: An Embodied Multimodal Language Model},
  author={Driess, Danny and Xia, Fei and Sajjadi, Mehdi SM and Lynch, Corey and Chowdhery, Aakanksha and Ichter, Brian and Wahid, Ayzaan and Tompson, Jonathan and Vuong, Quan and Yu, Tianhe and others},
  booktitle={International Conference on Machine Learning},
  pages={8469--8488},
  year={2023},
  organization={PMLR}
}

@inproceedings{chen2025fogros2,
  title={Fogros2-plr: Probabilistic latency-reliability for cloud robotics},
  author={Chen, Kaiyuan and Tian, Nan and Juette, Christian and Qiu, Tianshuang and Ren, Liu and Kubiatowicz, John and Goldberg, Ken},
  booktitle={2025 IEEE International Conference on Robotics and Automation (ICRA)},
  pages={16290--16297},
  year={2025},
  organization={IEEE}
}

@inproceedings{han2024pantheon,
  title={Pantheon: Preemptible multi-dnn inference on mobile edge gpus},
  author={Han, Lixiang and Zhou, Zimu and Li, Zhenjiang},
  booktitle={Proceedings of the 22nd Annual International Conference on Mobile Systems, Applications and Services},
  pages={465--478},
  year={2024}
}

@inproceedings{yi2020heimdall,
  title={Heimdall: mobile GPU coordination platform for augmented reality applications},
  author={Yi, Juheon and Lee, Youngki},
  booktitle={Proceedings of the 26th Annual International Conference on Mobile Computing and Networking},
  pages={1--14},
  year={2020}
}

@inproceedings{yi2020eagleeye,
  title={EagleEye: Wearable camera-based person identification in crowded urban spaces},
  author={Yi, Juheon and Choi, Sunghyun and Lee, Youngki},
  booktitle={Proceedings of the 26th Annual International Conference on Mobile Computing and Networking},
  pages={1--14},
  year={2020}
}

@inproceedings{jeong2022band,
  title={Band: coordinated multi-dnn inference on heterogeneous mobile processors},
  author={Jeong, Joo Seong and Lee, Jingyu and Kim, Donghyun and Jeon, Changmin and Jeong, Changjin and Lee, Youngki and Chun, Byung-Gon},
  booktitle={Proceedings of the 20th Annual International Conference on Mobile Systems, Applications and Services},
  pages={235--247},
  year={2022}
}

@article{yang2024active,
  title={Active Visual Perception Enhancement Method Based on Deep Reinforcement Learning},
  author={Yang, Zhonglin and Fang, Hao and Liu, Huanyu and Li, Junbao and Jiang, Yutong and Zhu, Mengqi},
  journal={Electronics},
  volume={13},
  number={9},
  pages={1654},
  year={2024},
  publisher={MDPI}
}

@inproceedings{lee2024learning,
  title={Learning to control camera exposure via reinforcement learning},
  author={Lee, Kyunghyun and Shin, Ukcheol and Lee, Byeong-Uk},
  booktitle={Proceedings of the IEEE/CVF Conference on Computer Vision and Pattern Recognition},
  pages={2975--2983},
  year={2024}
}

@article{von1993first,
  title={First Draft of a Report on the EDVAC},
  author={Von Neumann, John},
  journal={IEEE Annals of the History of Computing},
  volume={15},
  number={4},
  pages={27--75},
  year={1993},
  publisher={IEEE}
}

@inproceedings{bruyninckx2001open,
  title={Open robot control software: the OROCOS project},
  author={Bruyninckx, Herman},
  booktitle={Proceedings 2001 ICRA. IEEE international conference on robotics and automation (Cat. No. 01CH37164)},
  volume={3},
  pages={2523--2528},
  year={2001},
  organization={IEEE}
}

@article{lisberger2015visual,
  title={Visual guidance of smooth pursuit eye movements},
  author={Lisberger, Stephen G},
  journal={Annual review of vision science},
  volume={1},
  number={1},
  pages={447--468},
  year={2015},
  publisher={Annual Reviews}
}

@article{hopp2004characteristics,
  title={The characteristics and neuronal substrate of saccadic eye movement plasticity},
  author={Hopp, J Johanna and Fuchs, Albert F},
  journal={Progress in neurobiology},
  volume={72},
  number={1},
  pages={27--53},
  year={2004},
  publisher={Elsevier}
}

@inproceedings{sandler2018mobilenetv2,
  title={Mobilenetv2: Inverted residuals and linear bottlenecks},
  author={Sandler, Mark and Howard, Andrew and Zhu, Menglong and Zhmoginov, Andrey and Chen, Liang-Chieh},
  booktitle={Proceedings of the IEEE conference on computer vision and pattern recognition},
  pages={4510--4520},
  year={2018}
}

@article{comanici2025gemini,
  title={Gemini 2.5: Pushing the frontier with advanced reasoning, multimodality, long context, and next generation agentic capabilities},
  author={Comanici, Gheorghe and Bieber, Eric and Schaekermann, Mike and Pasupat, Ice and Sachdeva, Noveen and Dhillon, Inderjit and Blistein, Marcel and Ram, Ori and Zhang, Dan and Rosen, Evan and others},
  journal={arXiv preprint arXiv:2507.06261},
  year={2025}
}

@article{ashmore2008cochlear,
  title={Cochlear outer hair cell motility},
  author={Ashmore, Jonathan},
  journal={Physiological reviews},
  volume={88},
  number={1},
  pages={173--210},
  year={2008},
  publisher={American Physiological Society}
}

@book{moller2012hearing,
  title={Hearing: anatomy, physiology, and disorders of the auditory system},
  author={M{\o}ller, Aage R},
  year={2012},
  publisher={Plural Publishing}
}

@article{lederman2009haptic,
  title={Haptic perception: A tutorial},
  author={Lederman, Susan J and Klatzky, Roberta L},
  journal={Attention, Perception, \& Psychophysics},
  volume={71},
  number={7},
  pages={1439--1459},
  year={2009},
  publisher={Springer}
}

@article{johansson2009coding,
  title={Coding and use of tactile signals from the fingertips in object manipulation tasks},
  author={Johansson, Roland S and Flanagan, J Randall},
  journal={Nature Reviews Neuroscience},
  volume={10},
  number={5},
  pages={345--359},
  year={2009},
  publisher={Nature Publishing Group UK London}
}

@article{skljarevski2002nociceptive,
  title={The nociceptive flexion reflex in humans--review article},
  author={Skljarevski, V and Ramadan, NM},
  journal={Pain},
  volume={96},
  number={1-2},
  pages={3--8},
  year={2002},
  publisher={Elsevier}
}

@inproceedings{gong21b_interspeech,
  title     = {AST: Audio Spectrogram Transformer},
  author    = {Yuan Gong and Yu-An Chung and James Glass},
  year      = {2021},
  booktitle = {Interspeech 2021},
  pages     = {571--575},
  doi       = {10.21437/Interspeech.2021-698},
  issn      = {2958-1796},
}

@inproceedings{pham2019vdsn,
  title     = {Very Deep Self-Attention Networks for End-to-End Speech Recognition},
  author    = {Ngoc-Quan Pham and Thai Son Nguyen and Jan Niehues and Markus M{\"u}ller and Alexander H. Waibel},
  booktitle = {Proc.\ Interspeech},
  year      = {2019},
}

@article{chen2024tactile,
  title={Tactile-GAT: tactile graph attention networks for robot tactile perception classification},
  author={Chen, Lun and Zhu, Yingzhao and Li, Man},
  journal={Scientific Reports},
  volume={14},
  number={1},
  pages={27543},
  year={2024},
  publisher={Nature Publishing Group UK London}
}

@inproceedings{fan2022graph,
  title={Graph neural networks for interpretable tactile sensing},
  author={Fan, Wen and Bo, Hongbo and Lin, Yijiong and Xing, Yifan and Liu, Weiru and Lepora, Nathan and Zhang, Dandan},
  booktitle={2022 27th International Conference on Automation and Computing (ICAC)},
  pages={1--6},
  year={2022},
  organization={IEEE}
}

@inproceedings{tan2019efficientnet,
  title={Efficientnet: Rethinking model scaling for convolutional neural networks},
  author={Tan, Mingxing and Le, Quoc},
  booktitle={International conference on machine learning},
  pages={6105--6114},
  year={2019},
  organization={PMLR}
}

@inproceedings{howard2019searching,
  title={Searching for mobilenetv3},
  author={Howard, Andrew and Sandler, Mark and Chu, Grace and Chen, Liang-Chieh and Chen, Bo and Tan, Mingxing and Wang, Weijun and Zhu, Yukun arrows and Pang, Ruoming and Vasudevan, Vijay and others},
  booktitle={Proceedings of the IEEE/CVF international conference on computer vision},
  pages={1314--1324},
  year={2019}
}

@inproceedings{pech2000diatom,
  title={Diatom autofocusing in brightfield microscopy: a comparative study},
  author={Pech-Pacheco, Jos{\'e} Luis and Crist{\'o}bal, Gabriel and Chamorro-Martinez, Jes{\'u}s and Fern{\'a}ndez-Valdivia, Joaqu{\'\i}n},
  booktitle={Proceedings 15th International Conference on Pattern Recognition. ICPR-2000},
  volume={3},
  pages={314--317},
  year={2000},
  organization={IEEE}
}

@article{mink1996basal,
  title={The basal ganglia: focused selection and inhibition of competing motor programs},
  author={Mink, Jonathan W},
  journal={Progress in neurobiology},
  volume={50},
  number={4},
  pages={381--425},
  year={1996},
  publisher={Elsevier}
}

@article{mink2018basal,
  title={Basal ganglia mechanisms in action selection, plasticity, and dystonia},
  author={Mink, Jonathan W},
  journal={European Journal of Paediatric Neurology},
  volume={22},
  number={2},
  pages={225--229},
  year={2018},
  publisher={Elsevier}
}

@article{mcclelland1995there,
  title={Why there are complementary learning systems in the hippocampus and neocortex: insights from the successes and failures of connectionist models of learning and memory.},
  author={McClelland, James L and McNaughton, Bruce L and O'Reilly, Randall C},
  journal={Psychological review},
  volume={102},
  number={3},
  pages={419},
  year={1995},
  publisher={American Psychological Association}
}

@article{cole2013multi,
  title={Multi-task connectivity reveals flexible hubs for adaptive task control},
  author={Cole, Michael W and Reynolds, Jeremy R and Power, Jonathan D and Repovs, Grega and Anticevic, Alan and Braver, Todd S},
  journal={Nature neuroscience},
  volume={16},
  number={9},
  pages={1348--1355},
  year={2013},
  publisher={Nature Publishing Group US New York}
}

@article{duncan2010multiple,
  title={The multiple-demand (MD) system of the primate brain: mental programs for intelligent behaviour},
  author={Duncan, John},
  journal={Trends in cognitive sciences},
  volume={14},
  number={4},
  pages={172--179},
  year={2010},
  publisher={Elsevier}
}










\end{document}